\documentclass[11pt,a4paper,final, dvipsnames]{amsart}

\input xy
\xyoption{all}

\DeclareMathAlphabet{\mathpzc}{OT1}{pzc}{m}{it}

\usepackage{amsfonts,amsmath,amssymb,indentfirst,mathrsfs,amscd}
\usepackage[mathscr]{eucal}
\usepackage{graphicx}
\usepackage{nicefrac}
\usepackage{amsthm}
\usepackage{tikz-cd}

\usepackage{color}
\usepackage[utf8]{inputenc}
\usepackage[spanish, english]{babel}
\usepackage{booktabs}
\usepackage{multirow}
\usepackage{verbatim}
\usepackage{mathtools}
\usepackage[symbol]{footmisc}

\usepackage[smaller,nolist]{acronym}
\usepackage{cleveref}
\usepackage{subcaption}
\usepackage{mathtools}
\usepackage{amsmath}

\usepackage[symbol]{footmisc}

\usepackage{calrsfs}
\DeclareMathAlphabet{ \mathcal}{OMS}{zplm}{m}{n}

\usepackage[all]{xy}
\usepackage{graphicx}
\usepackage{stmaryrd}
\usepackage{enumitem}

\usepackage{empheq}
\usepackage{color}
\usepackage{xcolor}
\usepackage{dsfont}

\begin{acronym}
    \acro{DFT}{Discrete Fourier Transform}
    \acro{FFT}{Fast Fourier Transform}
    \acro{AGD}{Alternating Gradient Descend}
    \acro{RS}{Recommender System}
    \acro{CF}{Collaborative Filtering}
    \acro{MF}{Matrix Factorization}
    \acro{MAE}{Mean Absolute Error}
    \acro{MSE}{Mean Squared Error}
    \acro{SGD}{Stochastic Gradient Descent}
    \acro{DeepMF}{Deep Matrix Factorization}
    \acro{PMF}{Probabilistic Matrix Factorization}
    \acro{NMF}{Non-negative Matrix Factorization}
    \acro{CF4J}{Collaborative Filtering for Java}
    \acro{ALS}{Alternating Least Squares}
    \acro{ML}{Machine Learning}
    \acro{AI}{Artificial Intelligence}
    \acro{DL}{Deep Learning}
    \acro{GAN}{Generative Adversarial Network}
    \acro{NN}{Neural Network}
    \acro{ODE}{Ordinary Differential Equation}
    \acro{WGAN}{Wasserstein Generative Adversarial Network}
\end{acronym}{}

\newcommand{\cD}{\mathcal{D}}

\newcommand{\cF}{\mathcal{F}}
\newcommand{\cG}{\mathcal{G}} 
\newcommand{\cN}{\mathcal{N}} 

\newcommand{\CC}{\mathbb{C}} 
\newcommand{\TT}{\mathbb{T}} 
\newcommand{\RR}{\mathbb{R}} 
\newcommand{\ZZ}{\mathbb{Z}} 
\newcommand{\NN}{\mathbb{N}} 

\renewcommand{\Re}{\textrm{Re}} 

\newcommand{\Hess}{\mathrm{Hess}} 
\newcommand{\N}{\mathcal{N}} 

\address{Escuela Técnica Superior de Ingeniería de Sistemas Informáticos, Politécnica de Madrid, 28031 Madrid, Spain.}
\email{angel.gonzalez.prieto@upm.es \textrm{(Á.G.-P.)}; a.mozo@upm.es \textrm{(A.M.)};\newline{} e.talavera@upm.es \textrm{(E.T.)}; sm.gomez@upm.es \textrm{(S.G.-C.)}}

\author[]{\'Angel Gonz\'alez-Prieto}
\author[]{Alberto Mozo}
\author[]{Edgar Talavera}
\author[]{Sandra G\'omez-Canaval}

\title[]{Dynamics of Fourier Modes\\ in Torus Generative Adversarial Networks}


\begin{document}

\newtheorem{Theorem}{Theorem}[section]
\newtheorem{Proposition}[Theorem]{Proposition}
\newtheorem{Lemma}[Theorem]{Lemma}
\newtheorem{cor}[Theorem]{Corollary}
\newtheorem{conjecture}{Conjecture}
\newtheorem*{theorem*}{Theorem}
\newtheorem*{question*}{Question}

\theoremstyle{definition}
\newtheorem{Definition}[Theorem]{Definition}
\newtheorem{ex}[Theorem]{Example}
\newtheorem{as}{Assumption}

\theoremstyle{remark}
\newtheorem{Remark}[Theorem]{Remark}

\begin{abstract}
Generative Adversarial Networks (GANs) are powerful Machine Learning models capable of generating fully synthetic samples of a desired phenomenon with a high resolution. Despite their success, the training process of a GAN is highly unstable and typically it is necessary to implement several accessory heuristics to the networks to reach an acceptable convergence of the model. In this paper, we introduce a novel method to analyze the convergence and stability in the training of Generative Adversarial Networks. For this purpose, we propose to decompose the objective function of the adversary min-max game defining a periodic GAN into its Fourier series. By studying the dynamics of the truncated Fourier series for the continuous Alternating Gradient Descend algorithm, we are able to approximate the real flow and to identify the main features of the convergence of the GAN. This approach is confirmed empirically by studying the training flow in a $2$-parametric GAN aiming to generate an unknown exponential distribution. As byproduct, we show that convergent orbits in GANs are small perturbations of periodic orbits so the Nash equillibria are spiral attractors. This theoretically justifies the slow and unstable training observed in GANs.
\end{abstract}
\null
\vspace{-1.1cm}
\maketitle

\vspace{-0.8cm}

\section{Introduction}

Since their very inception, \acp{GAN} have revolutionized the areas of Machine Learning and Deep Learning. They address very successfully one of the most outstanding problems in pattern recognition: given a collection of examples of a certain phenomenon that we want to replicate, construct a generative model able to create new completely synthetic instances following the same patterns as the original ones. Ideally, the goal would be to capture the underlying pattern so subtlety that no external critic would be able to distinguish between real samples and synthesized instances.

The proposal of Goodfellow et al.\ \cite{Goodfellow:2014} is to confront two neural networks, in an adversary game, to solve this problem. More precisely, the proposal was to consider a neural network $G$ playing the role of a generator agent, and a network $D$ acting as discriminator. The discriminator $D$ is trained to distinguish as accurately as possible between real samples and fake/synthetic samples. On the other hand, $G$ aims to generate synthetic instances of high quality in such a way that $D$ is barely able to distinguish from real data. The two networks are, thus, in effective competition. When, as byproduct of this competition, the agents reach an optimal point we get a generator able to generate almost indistinguishable synthetic samples as well as a discriminator very proficient in classifying real and fake instances.

The way in which these networks are trained to reach this optimal point is through a common objective function. Explicitly, in \cite{Goodfellow:2014} it is proposed to consider the function
\begin{equation*}
    \mathcal{F}({\theta_D},{\theta_G}) = \mathbb{E}_\Omega \log\left[D_{\theta_D}(X)\right] + \mathbb{E}_\Lambda \log\left[1-D_{\theta_D}(G_{\theta_G})\right],
\end{equation*}
where $\theta_D$ are the inner weights of $D$, $\theta_G$ the weights of $G$, $\Omega$ is the probability space of the real data and $\Lambda$ is the latent probability space from which $G$ samples noise to be transformed into synthetic instances. In this manner, $\cF$ is essentially the error that $D$ suffers in the classification problem between real and fake examples so $D$ tries to maximize it and $G$ to minimize it. Hence, it gives rise to a non-convex min-max game and the goal of the training process is to reach a Nash equilibrium of it.

Several training approaches have been proposed to reach these Nash equlibria but the most widely used method is the so-called \ac{AGD}. Roughly speaking, the idea is to, alternatively, train $D$ by tuning $\theta_D$ with cost function $\cF$ and weights $\theta_G$ fixed and, after a certain amount of epochs, to reverse the roles and to update $\theta_G$ with cost function $-\cF$ and weights $\theta_D$ fixed. This optimization procedure has led to astonishing results, particularly in the domain of image processing and generation. Using several architectures and sophisticated multi-level training, \acp{GAN} are able to generate images with such a high quality that a human eye is not capable to distinguish them from real images \cite{karras2017progressive}. 

Despite of these achievements, stability of the \ac{AGD} algorithm for \acp{GAN} is a major issue. In \cite{Nagarajan-Kolter}, the authors proved that the Nash equillibria for \acp{GAN} are locally stable provided that some ideal conditions on the optimality of the equillibria are fulfilled. Nevertheless, these conditions may be unfeasible, as shown in \cite{Mescheder-Geiger:convergence}, so actual convergence and stability is not guaranteed in real applications. In particular, one of the most challenging problems arising during the training of \acp{GAN} is the so-called mode collapse \cite{goodfellow2016nips}. This state is characterized by a generator that has degenerated into a network that is only able to generate a single synthetic sample (or a very small number of them) with almost no variation, and such that the discriminator confuses with a real sample (typically, because the synthetic sample is actually very close to a real one). In this state, the system is no longer a generative model, but simply a copier of real data.

Furthermore, by construction, neural network-based \acp{GAN} have some intrinsic constraints in their expressivity that lead to very unrealistic synthetic samples in context far from image generation. For instance, neural networks produce a smooth output function, which provokes that \acp{GAN} have lots of difficulties to deal with the generation of real samples drawn from a discrete distribution (e.g.\ according to an exponential distribution) \cite{kusner2016gans}, or with some drastic semantic restrictions (e.g.\ non-negative values for counters) \cite{diesendruck2019importance}. These scenarios do not typically appear in image generation, but are common in other domains like data augmentation for Machine Learning \cite{antoniou2017data}. These problems lead to additional inconveniences for stable convergence and usually give rise to highly unstable models that require a very handcrafted stopping criteria and optimization heuristics.

Multitude of works have been oriented towards a deeper understanding of the instability of the training of \acp{GAN} as well as to propose solutions. A thorough theoretical study of the sources of instability and their causes can be found in \cite{arjovsky2017towards}, and in \cite{arora2017generalization,arora2018gans} the authors analyze the real capability of the \ac{GAN} for learning the distribution both through a theoretical and an empirical approach. In addition, in order to mitigate the instability of the training in \cite{salimans2016improved} the authors propose a collection of heuristical methods, through variations of the standard backpropagation algorithm, that contribute to stabilize the training process of \acp{GAN}. Moreover, in \cite{roth2017stabilizing} the use of regularization procedures is proposed to speed up the convergence.

Other very active research line is to propose of alternative models for \acp{GAN} that guarantee a better convergence. It is well known that the key reason why \ac{GAN} should capture the original distribution is because they implicitly optimize the Jensen-Shannon divergence (JSD) between the real underlying distribution and the generated distribution of the synthetic data \cite{Goodfellow:2014}. In order to change this framework, in \cite{arjovsky2017wasserstein} the authors propose to modify the cost function in such a way that the new \ac{GAN} does not optimize JSD but an Earth-mover distance known as Wasserstein distance, giving rise to the celebrated \acp{WGAN}. In a similar vein, in \cite{nowozin2016f} it is proposed to use the $f$-divergence (a divergence in the spirit of the Kullback-Leibler divergence) as criterion for training \acp{GAN}. Even genetic algorithms have been used to stabilize the training process, as in \cite{wang2019evolutionary}, where the authors applied genetic programming to optimize the use of different adversarial training objectives and evolved a population of generators to adapt to the discriminator, which acts as the hostile environment driving evolution. Nevertheless, despite of all these efforts, no master method is currently available and hence assuring a fast, or even effective, convergence of \acp{GAN} is an open problem.

\vspace{0.2cm}
\noindent\textbf{Our contribution.} In this paper we propose a novel method to analyze the convergence of \acp{GAN} through Fourier analysis. Concretely, we propose to approximate the objective function $\cF$ by its Fourier series, truncated with enough precision that the local dynamics of $\cF$ can be understood by means of a trigonometric polynomial.

Recall that any function $\cF(\theta): \TT^n \to \CC$ defined on the $n$-dimensional torus $\TT^n = (S^1)^n$ (equivalently, an $n$-periodic function on $\RR^n$) can be decomposed into a series of complex exponential functions, known as its Fourier series
$$
\cF(\theta) = \sum_{\mathbf{m} \in \ZZ^n} \alpha_{\mathbf{m}} \, e^{2\pi i \mathbf{m} \cdot \theta},
$$
where the series is indexed by the so-called Fourier modes or frequencies, $\mathbf{m}$ defined on the rectangular lattice $\ZZ^n \subseteq  \RR^n$. In principle, the previous equality must be understood as a decomposition in the Hilbert space of square-integrable functions, $L^2(\TT^n)$. However, if $\cF$ has enough regularity, then the Fourier series on the right hand side also converges uniformly to the original function $\cF$. This implies that, taking enough Fourier modes, $\cF$ can be effectively approximated by a truncated Fourier series. Moreover, if $\cF$ is real-valued, expressing the complex exponential as a combination of sine and cosine functions, we obtain an approximation of $\cF$ by a trigonometric polynomial, $\Theta(\cF)$.

This approximation can be applied to the study of the convergence of \acp{GAN} as follows. The continuous version of the \ac{AGD} algorithm can the though as a path of weights, $(\theta_D(t), \theta_G(t))$, depending on the time parameter $t \in \RR$. In particular, $(\theta_D(0), \theta_G(0))$ are the initial random weights of the \ac{GAN} and $(\theta_D(t), \theta_G(t))$ determine the state of the networks after training for a time $t > 0$. In this manner, if we seek to increase $\cF(\theta_D, \theta_G)$ in the direction $\theta_D$, and to decrease it in the direction $\theta_G$, the \ac{AGD} gives rise to a system of \acp{ODE} given by
\begin{equation*}
    \left\{
        \begin{matrix}
            \theta_D' = \nabla_{D} \cF(\theta_D, \theta_G), \\
            \theta_G' = -\nabla_{G} \cF(\theta_D, \theta_G), \\
        \end{matrix}
    \right.
\end{equation*}
where $\theta_D'$ and $\theta_G'$ denote the derivatives of the functions $\theta_D(t)$ and $\theta_G(t)$ with respect to the time $t$. This flow aims to converge to a Nash equilibrium of the objective function $\cF$ of the \ac{GAN} and, for this reason, we will refer to it as the Nash flow.

However, in many interesting cases the function $\cF$ may be very involved and lacks of an analytic closed expression which would enable an explicit analysis (e.g.\ even in the toy example of Equation (\ref{eq:cost-fun-example}) the cost function is intractable analytically). To address this problem, we propose to approximate $\cF$ by its truncated Fourier series, $\Theta(\cF)$. In this way, at least locally, the dynamic of the original Nash flow can be read from the solutions to the simplified system
\begin{equation*}
    \left\{
        \begin{matrix}
            \theta_D' = \nabla_{D} \Theta(\cF)(\theta_D, \theta_G), \\
            \theta_G' = -\nabla_{G} \Theta(\cF)(\theta_D, \theta_G). \\
        \end{matrix}
    \right.
\end{equation*}

In order to analyze this system of \acp{ODE}, we propose a novel method focused on studying the dynamics of the Nash flow on Fourier basic functions and on subsequent further approximations. As we will see, for the Nash flow of a basic trigonometric function, the Nash equillibria are not attractors of the flow but centers, that is, they are surrounded by periodic functions that spin around the critical point. When we consider more Fourier modes in the Fourier expansion of $\cF$, these periodic orbits may break leading to spiral attractors or spiral repulsors. The conditions that bifurcate the centers into spiral sinks or sources can be given explicitly in terms of the combinatorics of the considered Fourier modes.

This provides a theoretical justification to the empirically observed instability of the \ac{GAN} training: the convergent orbits towards a Nash equilibrium are mere perturbations of periodic orbits, falling slowly and spirally to the optimal point. For this reason, small variations in the training hyper-parameters, like the learning rate, the number of epochs or the batch size may lead to very different dynamics, which confers to the training its characteristic instability. In addition, in this paper we will empirically evaluate this method against a \ac{GAN} that aims to generate samples according to an unknown exponential distribution. To facilitate the visualization, we consider a simple \ac{GAN}, with $1$-dimensional parameter spaces each network, in such a way that the Nash flow can be plotted as a planar path. We will show that the proposed approach allow us to understand the simplified dynamics of the \ac{GAN} and to extract qualitative information of the Nash flow.

It is worth mentioning that, in order to have a natural Fourier series, the considered objective function $\cF$ of the \ac{GAN} must be periodic. This may seem unrealistic in real-life \acp{GAN}, but this is actually not a very strong condition. Usually, seeking to prove theoretical results about the convergence of \acp{GAN}, many works force that $\cF$ has compact support (for instance, to assure that it is Lipschitz as in \acp{WGAN}). In practice, this is accomplish by clipping the output of the generator and discriminator functions for large inputs. This provokes that, artificially, the objective function turns into a periodic function and, thus, it can be studied through the method introduced in this paper. We expect that this work will open the door to new methods for analyzing and quantifying the convergence of \acp{GAN} by importing well-established techniques of harmonic analysis and dynamical systems on closed manifolds, as studied in global analysis.

The structure of this paper is as follows. In Section \ref{sec:GAN-dynamics} we review the theoretical fundamentals of \acp{GAN} and their associated objective function and training method. In Section \ref{sec:Morse} we sketch briefly some basic concepts of Morse theory, a very successful theory that allows us to relate analytic properties of the function to be optimized with the topological properties of the underlying space. In Section \ref{sec:nash-flow} we introduce the Nash flow and discuss some of the arising problems for its convergence. In Section \ref{sec:torus-gans} we introduce torus \acp{GAN} and, particularly, in Section \ref{sec:fourier-torus} we explain how to perform Fourier analysis on the torus. Section \ref{sec:dynamics-fourier} is devoted to the analysis of the Nash flow for truncated Fourier series both for basic function (Sections \ref{sec:nash-flow-single-var} and \ref{sec:nash-flow-fourier-basis}) and for more complicated combinations (Sections \ref{sec:nash-flow-simplified} and \ref{sec:nash-flow-truncated}). In addition, in Section \ref{sec:empirical} the empirical testing of this method is performed, with comparisons between the real dynamic and the predicted ideal dynamic. Finally, in Section \ref{sec:conclusions} we summarize some of the keys ideas of this paper and sketch some lines of future work.

\subsection*{Acknowledgements}

The authors thank David Fontecha and María del Mar González for their careful reading of this manuscript and for pointing out several typos in a previous version. This work was supported in part by the European Union’s Horizon 2020 Research and Innovation Programme under Grant 833685 (SPIDER).

\section{GANs dynamics}\label{sec:GAN-dynamics}

As introduced by Goodfellow in \cite{Goodfellow:2014}, a \ac{GAN} network is a competitive model in which two intelligent agents (typically two neural networks) compete to improve their performance and to generate very precise samples according to a given distribution.

To be precise, let $X: \Omega \to \mathbb{R}^d$ be a $d$-dimensional random vector, defined on a certain probability space $\Omega$. This random vector $X$ should be understood as a very complex phenomenon whose instances we would like to replicate. For this purpose, we consider two functions
$$
    D: \RR^d \times \Theta_D \to \RR, \quad G: \Lambda \times \Theta_G \to \RR^d,
$$
called the \emph{discriminator} and the \emph{generator}, respectively. Here, $\Lambda$ is a probability space, called the latent space, and $\Theta_D, \Theta_G$ are two given topological spaces. These functions should be seen as parametric families of functions $D_{\theta_D}: \RR^d \to \RR$ and $G_{\theta_G}: \Lambda \to \RR^d$, parametrized by $\theta_D \in \Theta_D$ and $\theta_G \in \Theta_G$.

The aim of the \ac{GAN} is to tune the parameters $\theta_D$ and $\theta_G$ is such a way that, given $x \in \mathbb{R}^d$, $D_{\theta_D}(x)$ intends to predict whether $x = X(\omega)$ for some $\omega \in \Omega$ or not i.e.\ whether $x$ is compatible with being a real instance or it is a fake datum. Observe that, throughout this paper, we will follow the convention that $D_{\theta_D}(x)$ is the probability of being a real instance, so $D_{\theta_D}(x)=1$ means that $D_{\theta_D}$ is sure that $x$ is real and $D_{\theta_D}(x) = 0$ means that $D_{\theta_D}$ is sure that $x$ is fake. On the other hand, the generative function, $G_{\theta_G}$, is a $d$-dimensional random vector that seeks to converge in distribution to the original distribution $X$. Typically, the probability space $\Lambda$ is $\mathbb{R}^l$ with a certain standard probability distribution $\lambda$, as the spherical normal distribution or a uniform distribution on the unit cube.

\begin{Remark}
In typical applications in Machine Learning, $\Omega$ is given by a finite set $\Omega = \left\{x_1, \ldots, x_N\right\}$, with $x_i \in \mathbb{R}^d$, and endowed with a discrete probability (typically, the uniform one) so $X$ is just the identity function. In customary applications of \acp{GAN}, we have that the instances $x_i$ are images, represented by their pixel map, so the objective of the \ac{GAN} is to generate new images as similar as possible to the ones in the dataset $\Omega$.
\end{Remark}

The competition appears because the agents $D$ and $G$ try to improve non-simultaneously satifactible objectives. On one hand, $D$ tries to improve its performance in the classification problem but, on the other hand, $G$ tries to generate as best results as possible to cheat $D$. To be precise, recall that perfect fit for the classification problem for $D_{\theta_D}$ is given by $D_{\theta_D}(x)=1$ if $x$ is an instance of $X$ and $D_{\theta_D}(x)=0$ if not. Hence, the $L^1$ error made by $D_{\theta_D}$ with respect to perfect classification is
$$
    \mathcal{E}(\theta_D, \theta_G) = \mathbb{E}_\Omega \left[1-D_{\theta_D}(X)\right] + \mathbb{E}_\Lambda \left[D_{\theta_D}(G_{\theta_G})\right] = 1 - \mathbb{E}_\Omega \left[D_{\theta_D}(X)\right] + \mathbb{E}_\Lambda \left[D_{\theta_D}(G_{\theta_G})\right],
$$
where $\mathbb{E}_{\Omega}$ and $\mathbb{E}_{\Lambda}$ denote the mathematical expectation on $\Omega$ and $\Lambda$, respectively. In this way, the objective of $D_{\theta_D}$ is to minimize $\mathcal{E}$ while the goal of $G_{\theta_G}$ is to maximize it. It is customary in the literature to consider as objective the function $1-\mathcal{E}$ and to weight the error with a certain smooth concave function $f: \mathbb{R} \to \mathbb{R}$. In this way, the final cost function is
\begin{equation}\label{eq:cost-fun}
    \mathcal{F}({\theta_D},{\theta_G}) = \mathbb{E}_\Omega f\left[D_{\theta_D}(X)\right] + \mathbb{E}_\Lambda f\left[-D_{\theta_D}(G_{\theta_G})\right] .
\end{equation}

\begin{Remark}
Typical choices for the weight function $f$ are $f(s) = - \log(1 + \exp(-s))$, as in the original paper of Goodfellow \cite{Goodfellow:2014}, or $f(s) = s$ as in the Wasserstein GAN \cite{Arjovsky-WGAN}. 
\end{Remark}

However, in sharp contrast with what is typical in Machine Learning, the aim of the \ac{GAN} is not to maximize/minimize $\cF$. The objectives of the $D$ and $G$ agents are opposed: while $D$ tries to maximize $\cF$, the generator tries to minimize it. In this vein, the objective of the \ac{GAN} is
\begin{align}\label{eq:min-max-game}
    \min_{{\theta_G}}\,\max_{{\theta_D}} \mathcal{F}({\theta_D},{\theta_G}) = \min_{{\theta_G}}\,\max_{{\theta_D}}\mathbb{E}_\Omega f\left[D_{\theta_D}(X)\right] + \mathbb{E}_\Lambda f\left[-D_{\theta_D}(G_{\theta_G})\right].
\end{align}

In the case that the latent space $\Lambda$ is naturally equipped with a topology (as in the case $\Lambda = (\RR^l, \lambda)$), in is customary to require that $\cF: \Theta_D \times \Theta_G \to \RR$ is a continuous function. In addition, in our case $\Theta_G$ and $\Theta_D$ will be differentiable manifolds, so we will require that both $D$ and $G$ are $C^2$ maps in both arguments and, thus, $\cF$ is a differentiable function on $\Theta_D \times \Theta_G$.

To be precise, algorithm proposed by Goodfellow \cite{Goodfellow:2014} suggests to freeze the internal weights of $G$ and to use it to generate a batch of fake examples from $\Lambda$. With this set of fake instances and another batch of real instances created using $X$ (i.e.\ sampling randomly from the dataset of real instances), we train $D$ to improve its accuracy in the classification problem with the usual backpropagation (i.e.\ gradient descent) method. Afterwards, we freeze the weights of $D$ and we sample a batch of latent data of $\Lambda$ (i.e.\ we randomly sample noise using the latent distribution) and we use it to train $G$ using gradient descent for $G$ with objective function $\theta_G \mapsto \mathbb{E}_{\Lambda}f(-D(G_{\theta_G}))$. Finally, we can alternate this process as many times as needed until we reach the desired results. Several metrics have been proposed to quantify this performance, specially regarding the domain of image generation, like Inception Score (IS) \cite{salimans2016improved}, Fr\'echet Inception Distance (FID) \cite{heusel2017gans} or perceptual similarity measures \cite{snell2017learning}. For a survey of these techniques, please refer to \cite{borji2019pros}.

\subsection{Review of Morse theory}\label{sec:Morse}

Let us suppose for a while that, instead of looking for solutions of (\ref{eq:min-max-game}) we were seeking to local maxima of $\cF$. In this situation, the standard approach in Machine Learning is to consider the Morse flow, also known as gradient ascent flow. For it, let us fix riemannian metrics on $\Theta_D$ and $\Theta_G$. Using them, we can compute the \emph{gradient} of $\cF$, $\nabla \cF = (\nabla_{D} \cF, \nabla_G \cF)$, where $\nabla_{D} \cF, \nabla_G \cF$ denote the gradient in the $\theta_D, \theta_G$ direction ,respectively. Then, the Morse flow is the differentiable flow on $\Theta_D \times \Theta_G$ generated by the vector field $\nabla \cF$. Explicitly, it is given by the system of \acp{ODE}
\begin{equation}\label{eq:morse-flow}
    \left\{
        \begin{matrix}
            \theta_D' = \nabla_{D} \cF(\theta_D, \theta_G), \\
            \theta_G' = \nabla_{G} \cF(\theta_D, \theta_G). \\
        \end{matrix}
    \right.
\end{equation}
This flow has been objective of very intense studies in the context of differentiable geometry and geometric topology. For instance, it is the crucial tool used in Smale's proof of the Poincar\'e conjecture in high dimension \cite{milnor2015lectures}, and has been successfully used to understand the topology of moduli spaces of solutions to highly non-linear Partial Differential Equations coming from theoretical physics \cite{atiyah1983yang}, among others.

Obviously, the critical points of the system (\ref{eq:morse-flow}) are exactly the \emph{critical points} of $\cF$, in the sense that the differential $d\cF|_{(\theta_D^0, \theta_G^0)} = 0$. In order to control the dynamics of this \ac{ODE} around a critical point, a key concept is the notion of index of a point.

\begin{Definition}
Let $(\theta_D^0, \theta_G^0)$ be a critical point of $\cF$. The \emph{Hessian} of $\cF$ at $(\theta_D^0, \theta_G^0)$ is the symmetric $2$-form $H\cF|_{\theta_D^0, \theta_G^0} \in \textrm{Sym}^2(T^*_{\theta_D^0} \Theta_D \oplus T^*_{\theta_G^0} \Theta_G)$ given by
$$
    \Hess(\cF)|_{\theta_D^0, \theta_G^0}(v,w) = w(\tilde{v}(\cF)),
$$
for $v \in T_{\theta_D^0} \Theta_D, w \in T_{\theta_G^0}\Theta_G$ and $\tilde{v}$ any extension of $v$ to an vector field in a small neighborhood of $(\theta_D^0, \theta_G^0)$.

The point $(\theta_D^0, \theta_G^0)$ is said to be \emph{non-degenerate} if $\Hess(\cF)|_{\theta_D^0, \theta_G^0}$ is non-degenerated as $2$-form. In that case, the \emph{index} of the point, denoted $\lambda(\theta_D^0, \theta_G^0)$, is the number of negative eigenvalues of $\Hess(\cF)|_{\theta_D^0, \theta_G^0}$. A function $\cF$ is said to be \emph{Morse} if all its critical points are non-degenerate.
\end{Definition}

More explicitly, let $\partial_D^1, \ldots, \partial^{d_D}_D$ be a basis of $T_{\theta_D^0} \Theta_D$ and $\partial^1_G, \ldots, \partial^{d_G}_G$ be a basis of $T_{\theta_G^0} \Theta_G$, where $d_D$ and $d_G$ are the dimensions of $\Theta_D$ and $\Theta_G$ respectively. Then, Hessian is the matrix of second derivatives
$$
    \Hess(\cF) = \begin{pmatrix}
    \frac{\partial^2 \cF}{\partial \theta_D^i \partial \theta_D^j} & \frac{\partial^2 \cF}{\partial \theta_D^i \partial \theta_G^j} \\
    \frac{\partial^2 \cF}{\partial \theta_G^i \partial \theta_D^j} & \frac{\partial^2 \cF}{\partial \theta_G^i \partial \theta_G^j}
    \end{pmatrix}
$$

If $\Theta_D$ and $\Theta_G$ are compact, Morse functions are known to form a dense open set of the space of continuous functions on $\Theta_D \times \Theta_D$ \cite{milnor2015lectures}. Moreover, the critical points of a Morse function are isolated, in the sense that there exists an open neighborhood of each critical point that contains only that critical point. Indeed, the stability of a critical point $(\theta_D, \theta_G)$ is fully determined by its index. Then, $(\theta_D, \theta_G)$ is a sink in a hypersurface of dimension $\lambda(\theta_D, \theta_G)$, while it is a source in a hypersurface of dimension $d_Dd_G - \lambda(\theta_D, \theta_G)$. In particular, the only sinks of the Morse flow are precisely the local maxima of $\cF$, in which $\Hess(\cF)$ is negative-define and, thus, $\lambda(\theta_D, \theta_G)=d_Dd_G$.

Another important fact that we will use is the following topological interpretation of the indices, known as the Poincar\'e-Hopf theorem. It claims that, if $\Theta_D$ and $\Theta_G$ are compact then
\begin{align}\label{eq:crit-Euler}
    \sum_{(\theta_D, \theta_G) \in \textrm{Crit}(\cF)} (-1)^{\lambda(\theta_D, \theta_G)} = \chi(\Theta_D \times \Theta_G) = \chi(\Theta_D) \chi(\Theta_G).
\end{align}
Here, $\textrm{Crit}(\cF)$ denotes the (finite) set of critical points of $\cF$ and $\chi$ is the Euler characteristic of the space.

\subsection{The Nash flow}\label{sec:nash-flow}

Now, let us come back to our optimization problem (\ref{eq:min-max-game}). Despite of the simplicity of the formulation of the cost function, this problem is very far from being trivial. The best scenario would be to obtain a so-called Nash equilibrium.

\begin{Definition}
Let $\cF: \Theta_D \times \Theta_G \to \RR$ be a differentiable function. A point $(\theta_D^0, \theta_G^0) \in  \Theta_D \times \Theta_G$ is said to be a \emph{Nash equilibrium} if:
\begin{itemize}
    \item The function $\theta_D \mapsto \cF(\theta_D, \theta_G^0)$ has a maximum at $\theta_D^0$.
    \item The function $\theta_G \mapsto \cF(\theta_D^0, \theta_G)$ has a minimum at $\theta_G^0$.
\end{itemize}
\end{Definition}

\begin{Remark}
A Nash equilibrium is in particular a {critical point} of $\cF$.
\end{Remark}

In this vein, it is natural to consider an analogous differentiable flow to (\ref{eq:morse-flow}) but converging to Nash equilibria. For this purpose, fix riemannian metrics on $\Theta_D$ and $\Theta_G$ as above and consider the gradient $\nabla \cF = (\nabla_D \cF, \nabla_G \cF)$. Now, we twist the gradient to consider the \emph{Nash vector field}
$$
    \N(\cF) = (\nabla_D \cF, -\nabla_G \cF).
$$

\begin{Definition}
The \emph{Nash flow} is the differentiable flow on $\Theta_D \times \Theta_G$ generated by the Nash vector field $\N(\cF)$. Explicitly, it the the system of \acp{ODE}
\begin{equation}\label{eq:nash-flow}
    \left\{
        \begin{matrix}
            \theta_D' = \nabla_{D} \cF(\theta_D, \theta_G), \\
            \theta_G' = -\nabla_{G} \cF(\theta_D, \theta_G). \\
        \end{matrix}
    \right.
\end{equation}
\end{Definition}

This flow (or, more precisely, the associated discrete-time version known as the \ac{AGD} flow) has been intensively used for training \acp{GAN} from their very inception. Already in Goodfellow's seminar paper \cite{Goodfellow:2014}, this flow is proposed as a method for seeking to a Nash equilibriums of the game (\ref{eq:min-max-game}).

To understand the dynamics of the Nash flow, let us study it around a critical point. Working in a local chart around a critical point, with an adapted basis $\partial_D^1, \ldots, \partial^{d_D}_D, \partial^1_G, \ldots, \partial^{d_G}_G$ of $T_{\theta_D^0} \Theta_D \oplus T_{\theta_G^0} \Theta_G$, have that the differential of the Nash vector field is the Nash Hessian
$$
    \N\Hess(\cF) = \left(\N(\cF)\right)_* = \begin{pmatrix}
    \frac{\partial^2 \cF}{\partial \theta_D^i \partial \theta_D^j} & \frac{\partial^2 \cF}{\partial \theta_D^i \partial \theta_G^j} \\
    -\frac{\partial^2 \cF}{\partial \theta_G^i \partial \theta_D^j} & -\frac{\partial^2 \cF}{\partial \theta_G^i \partial \theta_G^j}
    \end{pmatrix}
$$
In this manner, in a small neighborhood of a critical point $(\theta_D^0,\theta_G^0) \in \Theta_D \times \Theta_G$ of $\cF$ (in particular, around a Nash equilibrium), the dynamics are determined by the linearized version
\begin{equation*}
    \left\{
        \begin{pmatrix}
            \theta_D' \\
            \theta_G'
        \end{pmatrix} = \left.\begin{pmatrix}
    \frac{\partial^2 \cF}{\partial \theta_D^i \partial \theta_D^j} & \frac{\partial^2 \cF}{\partial \theta_D^i \partial \theta_G^j} \\
    -\frac{\partial^2 \cF}{\partial \theta_G^i \partial \theta_D^j} & -\frac{\partial^2 \cF}{\partial \theta_G^i \partial \theta_G^j}
    \end{pmatrix}\right|_{(\theta_D^0,\theta_G^0)}        \begin{pmatrix}
            \theta_D \\
            \theta_G
        \end{pmatrix}
    \right.
\end{equation*}
However, in sharp contrast with the Morse flow, even if $\cF$ has non-degenerate critical points, it may happen that the Nash equilibria are not attractors. For instance, if the Nash Hessian has vanishing diagonal (as in Section \ref{sec:nash-flow-fourier-basis}), then periodic orbits arise around the critical point and the flow is non-convergent.

Nonetheless, this behavior can be controlled. Suppose for simplicity that $d_D=d_G=1$ (higher dimensional scenarios can be treated analogously by splitting the tangent space). In that case, the eigenvalues of $\N\Hess(\cF)$ are either both real or complex conjugated.
\begin{itemize}
    \item If the eigenvalues are real, around a Nash equilibrium both eigenvalues must be non-negative since in the usual Hessian they have different signs. Hence, the Nash equilibrium is a non-repulsor of the Nash flow. Moreover, if $\cF$ is Morse, then its eigenvalues do not vanish and, thus, the Nash equilibrium is an attractor.
    \item If the eigenvalues are complex conjugated, say $\lambda,\overline{\lambda}\in \CC$, then the dynamic is controlled by the real part of $\lambda$, $\Re(\lambda)$. There is an invariant way of computing this quantity as through the trace of $\N\Hess(\cF)$ since
    $$
        2\Re(\lambda) = \lambda + \overline{\lambda} = \textrm{tr}\left( \N\Hess(\cF)\right) = \frac{\partial^2 \cF}{\partial \theta_D^2} - \frac{\partial^2 \cF}{\partial \theta_G^2}.
    $$
    Observe that this is nothing but the wave operator acting on $\cF$. In the case that this trace is negative, the critical point is an attractor with spiral dynamic; if it is positive, it is a repulsor; and if it vanishes, it is a center with surrounding periodic orbits.
\end{itemize}

It is worth mentioning that, in the case of \acp{GAN}, the function $\cF$ of (\ref{eq:min-max-game}) to be optimized does not define a convex-concave game so, in general, the convergence of the usual training methods through Nash flow is not guaranteed \cite{Nagarajan-Kolter}. Under some ideal assumptions on the behaviour of the game around the Nash equilibrium points, in \cite{Nagarajan-Kolter} the authors proved that the Nash flow is locally asymptotically stable. However, the hypotheses needed to apply this result are quite strong and seem to be unfeasible in practice. For instance, in \cite{Mescheder-Geiger:convergence}, the authors show an example of a very simple \ac{GAN}, the so-called Dirac \ac{GAN}, for which the usual gradient descend does not converge.

\section{Torus GANs}\label{sec:torus-gans}

From now on, let us focus on a very particular case of \ac{GAN}, that we shall call a \emph{torus \ac{GAN}}. Let us denote
$$
    \TT^n = \underbrace{S^1 \times \ldots S^1}_{n \textrm{ times}}
$$
the $n$-dimensional torus. Then, we will take as parameter spaces
$\Theta_D = \TT^{d_D}$ and $\Theta_G = \TT^{d_G}$. In this way, the cost functional becomes a function
$$
    \cF: \TT^{d_D} \times \TT^{d_G} = \TT^{d_D + d_G} \to \RR.
$$

\begin{Remark}
This particular choice is not as arbitrary as it may seem at a first sight. In the end, a torus \ac{GAN} is any \ac{GAN} in which the generator and discriminator are periodic functions on their parameters $\theta_D$ and $\theta_G$ for some large enough period. In standard neural network-based \acp{GAN} it is customary to clip the output of the neural network in order to prevent the internal weights to become arbitrary large. This is particularly important specially Wasserstein \acp{GAN}, where the objective function is required to be Lipschitz and this is achieved by forcing the cost function to have compact support. In this way, after clipping, both the generator and the discriminator agents are periodic functions and, thus, they define a torus \ac{GAN}.
\end{Remark}

Working on the torus has important consequences to the dynamics the Morse flow. Some of them are the following:
\begin{itemize}
    \item Divergent orbits are not allowed. Since $\TT^n$ is compact, standard results of prologability of solutions for short-time show that the orbits of any vector flow cannot blow-up. Intuitively, they cannot escape by tending to infinity. In particular, if $\cF$ is a Morse function, all the orbits in the Morse flow must converge to a critical point. This is a consequence of the fact that, along a non-constant orbit of the Morse flow , the function $\cF$ is strictly increasing since
    $$
        \frac{d}{dt}\cF(\theta_D, \theta_G) = d\cF(\theta_D', \theta_G') = d\cF(\nabla \cF) = ||\nabla \cF||^2 > 0.
    $$
    Thus, since $\cF$ is bounded, the flow is forced to converge to a constant orbit, that is, to a critical point of $\cF$. This prevents the appearance of periodic orbits in the Morse flow. In the Nash flow, this may no longer hold and periodic orbits may arise (as in Section \ref{sec:nash-flow-fourier-basis}).
    \item Topological restrictions. The Euler characteristic of $\TT^n$ is $\chi(\TT^n) = \chi(S^1)^n = 0$. Hence, equation (\ref{eq:crit-Euler}) implies that
    $$
        \sum_{(\theta_D, \theta_G) \in \textrm{Crit}(\cF)} = 0.
    $$
    In other word, there is the same number of critical points of even index than of odd index. In particular, if $d_D = d_G = 1$, there are as many saddle points (which are points of index $1$) as maxima and minima (which are points of index $2$ or $0$). 
\end{itemize}

\subsection{Fourier analysis in the torus}\label{sec:fourier-torus}

In order to understand the cost function $\cF$ of a torus \ac{GAN}, we shall apply techniques of harmonic analysis to it. We will suppose that the reader is familiar with basic notions of Fourier and harmonic analysis, like Hilbert spaces and orthogonal  Schauder basis on them. Otherwise, please refer to \cite{rudin2006real}.

Let us see $\TT^n = \RR^n/\ZZ^n$ so that functions on $\TT^n$ are $n$-periodic functions on the unit square. Recall that a fundamental result of Fourier analysis is that the space $L^2(\TT^n)$ of complex-valued square-integrable functions on $\TT^n$ is a Hilbert space with product given by
$$
    \langle \cF, \cG\rangle = \int_{\TT^n} \cF(\theta) \overline{\cG(\theta)} \, d\theta.
$$
Moreover, this space is spanned by the orthonormal basis of functions
$$
    e_{\mathbf{m}}(\theta) = e^{2\pi i \mathbf{m} \cdot \theta},
$$
where $\mathbf{m} = (m_1, \ldots, m_n) \in \ZZ^n$, $\theta = (\theta_1, \ldots, \theta_n) \in \TT^n$ and $\mathbf{m} \cdot \theta = m_1 \theta_1 + \ldots + m_n \theta_n$ is the standard inner product. In other words, any $\cF \in L^2(\TT^n)$ can be uniquely written as a sum
$$
    \cF(\theta) = \sum_{\mathbf{m} \in \ZZ^n} \alpha_{\mathbf{m}}\, e_{\mathbf{m}}(\theta) =  \sum_{\mathbf{m} \in \ZZ^n} \alpha_{\mathbf{m}} \, e^{2\pi i \mathbf{m} \cdot \theta},
$$
in the sense that this sum is convergent in $L^2(\TT^n)$ and converges to $\cF$. This expression is referred to as the \emph{Fourier series} of $\cF$. The coefficients $\alpha_{\mathbf{m}}$ are called the \emph{Fourier coefficients} or the \emph{Fourier modes} of $\cF$. Using the orthogonality of the functions $e_{\mathbf{m}}(\theta)$, they can be obtained as
$$
    \alpha_{\mathbf{m}} = \langle \cF, e_{\mathbf{m}}(\theta) \rangle = \int_{\TT^n} \cF(\theta) e^{-2\pi i \mathbf{m} \cdot \theta} \,d\theta.
$$

In principle, the convergence of the Fourier series to $\cF$ is only in the $L^2$ sense (c.f.\ \cite{du1873ueber} for a Fourier series of a continuous function not converging pointwise everywhere, or \cite{kolmogorov1926series} for an everywhere divergent Fourier series of a $L^1$ function). However, if $\cF$ is $C^1$, since we are working on a compact space, it is automatically H\"older and, thus, its Fourier series converges uniformly \cite{zygmund2002trigonometric}. This means that, for every $\epsilon > 0$
$$
    \left|\left|\cF - \sum_{m_i = -N}^N \alpha_{\mathbf{m}} \, e_{\mathbf{m}}\right|\right|_\infty = \sup_{\theta \in \TT^n} \left|\cF(\theta) - \sum_{m_i = -N}^N \alpha_{\mathbf{m}} \, e^{2\pi i \mathbf{m} \cdot \theta}\right| < \epsilon,
$$
for all $N$ large enough. Similar approximations can be obtained for the $k$ first derivatives of $\cF$ if it has enough regularity (concretely, if it is $C^{k+1}$).

This approximation is very useful for estimating the associated flow. Recall that, using Gronwall inequality \cite{gronwall1919note}, if $X, Y$ are two Lipschitz vector fields, then there exists a constant $M > 0$ such that their associated flows $\theta(t)$ and $\vartheta(t)$ satisfy
$$
    |\theta(t) - \vartheta(t)| \leq  \frac{e^{Mt}-1}{M} ||X-Y||_{\infty}
$$
for all $t$. In other words, for medium-times, the flow of $X$ may be approximated through the flow of $Y$.

\begin{Remark}
The previous estimation implies that, locally, the dynamics of the flows $\theta(t)$ and $\vartheta(t)$ are similar. In particular, this is useful for analyzing convergence around critical points. Nevertheless, the global dynamics of $\theta(t)$ and $\vartheta(t)$ may be quite different, say, they may have different numbers of critical points.
\end{Remark}

In our context, this idea can be exploited as follows. Let us denote by
$$
    \Theta_N(\cF) =  \sum_{m_i = -N}^N \alpha_{\mathbf{m}} \, e_{\mathbf{m}}
$$
the truncated Fourier series of $\cF$. If $\cF$ is $C^2$, then $\nabla \cF$ and $\nabla \Theta_N(\cF)$ are close vector fields and, thus
$$
    |\theta(t) - \theta_N(t)| \leq  \frac{e^{Mt}-1}{M} ||\nabla \cF-\nabla \Theta_N(\cF)||_{\infty} \leq \epsilon(e^{Mt}-1) 
$$
for $N$ large enough, where $\theta(t)$ is the Morse flow for $\cF$ and $\theta_N(t)$ is the Morse flow for $\Theta_N(\cF)$. Working verbatim with the Nash vector fields we obtain similar estimates for the solutions of the Nash flow.

\section{Dynamics of Fourier basis}\label{sec:dynamics-fourier}

In this section, we focus on the Nash flow of truncated approximations of Fourier series of a $C^2$ function $\cF$. As we mentioned above, these solutions approximate quite well the real Nash flow of $\cF$ for short times (particularly, around critical points).

For the sake of simplicity, in this section we shall focus on the $2$-dimensional case in which $d_D = d_G = 1$ so that $\cF = \cF(\theta_1, \theta_2)$ is a function
$$
    \cF: \TT^2 \to \RR.
$$
Moreover, we will truncate the Fourier series at level $N=2$. Similar arguments can be carried out for higher dimension and more accurate precision of the Fourier series with similar results, but the calculations become more involved.

First of all, let us re-write the Fourier series of $\cF$ as a trigonometric polynomial. Recall that the trigonometric functions can be obtained from the complex exponential as
$$
    \cos(2 \pi \theta) = \frac{e^{2 \pi i \theta} + e^{-2 \pi i \theta}}{2}, \quad \sin(2\pi \theta) = \frac{e^{2 \pi i \theta} - e^{-2 \pi i \theta}}{2i}.
$$

Since the function $\cF$ is real-valued, we can group the coefficients and to obtain a formula for the Fourier series in term of trigonometric functions as
\begin{align*}
    \cF(\theta_1, \theta_2) &= \sum_{m_1, m_2 = 0}^\infty a_{m_1, m_2}^{0,0}  \sin(2 \pi m_1 \theta_1)  \sin(2 \pi m_2\theta_2) +  \sum_{m_1, m_2 = 0}^\infty a_{m_1, m_2}^{0,1}  \sin(2 \pi m_1 \theta_1)  \cos(2 \pi m_2\theta_2) \\
     &\;\;\;+ \sum_{m_1, m_2 = 0}^\infty a_{m_1, m_2}^{1,0}  \cos(2 \pi m_1 \theta_1)  \sin(2 \pi m_2\theta_2) +  \sum_{m_1, m_2 = 0}^\infty a_{m_1, m_2}^{1,1}  \cos(2 \pi m_1 \theta_1)  \cos(2 \pi m_2\theta_2).
\end{align*}

The coefficients are real numbers that can be obtained as
\begin{align*}
    a_{m_1,m_2}^{0,0} &= \delta_{m_1,m_2} \langle \cF,  \sin(2 \pi m_1 \theta_1)  \sin(2 \pi m_2\theta_2)\rangle = \delta_{m_1,m_2}\int_{\TT^2}\cF(\theta_1,\theta_2) \sin(2 \pi m_1 \theta_1)  \sin(2 \pi m_2\theta_2) \,d\theta_1d\theta_2, \\
    a_{m_1,m_2}^{0,1} &= \delta_{m_1,m_2} \langle \cF,  \sin(2 \pi m_1 \theta_1)  \cos(2 \pi m_2\theta_2)\rangle = \delta_{m_1,m_2}\int_{\TT^2}\cF(\theta_1,\theta_2) \sin(2 \pi m_1 \theta_1)  \cos(2 \pi m_2\theta_2) \,d\theta_1d\theta_2, \\
    a_{m_1,m_2}^{1,0} &= \delta_{m_1,m_2} \langle \cF,  \cos(2 \pi m_1 \theta_1)  \sin(2 \pi m_2\theta_2)\rangle = \delta_{m_1,m_2}\int_{\TT^2}\cF(\theta_1,\theta_2) \cos(2 \pi m_1 \theta_1)  \sin(2 \pi m_2\theta_2) \,d\theta_1d\theta_2, \\
    a_{m_1,m_2}^{1,1} &= \delta_{m_1,m_2} \langle \cF,  \cos(2 \pi m_1 \theta_1)  \cos(2 \pi m_2\theta_2)\rangle = \delta_{m_1,m_2}\int_{\TT^2}\cF(\theta_1,\theta_2) \cos(2 \pi m_1 \theta_1)  \cos(2 \pi m_2\theta_2) \,d\theta_1d\theta_2,
\end{align*}
where $\delta_{m_1,m_2}$ is a coefficient that $\delta_{m_1, m_2} = 1$ if $m_1 = m_2 = 0$; $\delta_{m_1, m_2} = 2$ if $m_1 = 0$ and $m_2 > 0$, $m_1 > 0$ and $m_2 = 0$; and $\delta_{m_1, m_2} = 4$ $m_1, m_2 > 0$.

To shorten notation, from now on we shall denote
\begin{align*}
    \Lambda_{m_1, m_2}^{0,0}(\theta_1, \theta_2) &=  \sin(2 \pi m_1 \theta_1)  \sin(2 \pi m_2\theta_2),\quad&
    \Lambda_{m_1, m_2}^{0,1}(\theta_1, \theta_2) &=  \sin(2 \pi m_1 \theta_1)  \cos(2 \pi m_2\theta_2),\\
    \Lambda_{m_1, m_2}^{1,0}(\theta_1, \theta_2) &=  \cos(2 \pi m_1 \theta_1)  \sin(2 \pi m_2\theta_2),\quad&
    \Lambda_{m_1, m_2}^{1,1}(\theta_1, \theta_2) &=  \cos(2 \pi m_1 \theta_1)  \cos(2 \pi m_2\theta_2),
\end{align*}
This notation is particularly useful because, for any $\alpha, \beta \in \ZZ_2$
$$
     \frac{\partial}{\partial \theta_1}\Lambda_{m_1, m_2}^{\alpha,\beta} = (-1)^\alpha 2\pi m_1 \Lambda_{m_1, m_2}^{\alpha+1,\beta}, \quad \frac{\partial}{\partial \theta_2}\Lambda_{m_1, m_2}^{\alpha,\beta} = (-1)^\beta 2\pi m_2 \Lambda_{m_1, m_2}^{\alpha,\beta+1},
$$
where the sum is interpreted as sum in $\ZZ_2$.

From this expression of the Fourier series, we will approximate the dynamics of the Nash flow for $\cF$ by truncating the Fourier series. In particular, we sort the coefficients $a_{m_1,m_2}^{\alpha,\beta}$ by decreasing order of their absolute value. Looking only at the two largest coefficients, and normalizing so that the leading coefficient is $1$, we will consider the approximation to $\cF$
\begin{align}\label{eq:Fourierapprx}
    \Theta(\cF) = \Lambda_{m_1, m_2}^{\alpha,\beta} + \mu \Lambda_{n_1, n_2}^{\gamma,\delta},
\end{align}
where $\alpha, \beta, \gamma, \delta \in \ZZ_2$, $(m_1, m_2)$ are the leading Fourier modes and $(n_1, n_2)$ are the second largest modes, and $|\mu| < 1$.

\subsection{Nash flow for single variable Fourier basis}\label{sec:nash-flow-single-var}

From now on, we aim to analyze the Nash flow for a truncated Fourier series. As we will see in Section \ref{sec:empirical}, from it we can envisage the global dynamics of the Nash flow for the objective function of a \ac{GAN}.

First of all, let us consider the simplest Fourier modes, namely with $m_1 = 0$ or $m_2 = 0$. In this case, the dynamics is quite simple and, in most cases, can be pulled apart. In the case of $\Lambda_{0,0}^{\alpha, \beta}(\theta_1, \theta_2) \equiv 1$, the Nash flow equations amount to
\begin{equation*}
    \left\{
        \begin{matrix}
            \theta_1' = \frac{\partial}{\partial \theta_1} \Lambda_{0,0}^{\alpha, \beta}(\theta_1, \theta_2) = 0, \\
            \theta_2' = -\frac{\partial}{\partial \theta_2} \Lambda_{0,0}^{\alpha, \beta}(\theta_1, \theta_2) = 0. \\
        \end{matrix}
    \right.
\end{equation*}
Therefore, the solutions are constant orbits $(\theta_1(t), \theta_2(t))=(\theta_1^0, \theta_2^0)$ for some fixed $(\theta_1^0, \theta_2^0) \in \TT^2$. For this reason, it does not contribute to the dynamics.

For Fourier modes of the form $\Lambda_{m_1,0}^{0, \beta}(\theta_1, \theta_2) = \sin(2\pi m_1 \theta_1)$ or $\Lambda_{m_1,0}^{1, \beta}(\theta_1, \theta_2) = \cos(2\pi m_1 \theta_1)$ the situation is also very simple. Now, the Nash flow is given by
\begin{equation*}
    \left\{
        \begin{matrix*}[l]
            \theta_1' &= \frac{\partial}{\partial \theta_1}  \Lambda_{m_1,0}^{\alpha, \beta}(\theta_1, \theta_2) = 2\pi m_1 \Lambda_{m_1,0}^{\alpha+1, \beta}(\theta_1, \theta_2), \\
            \theta_2' &= -\frac{\partial}{\partial \theta_2}  \Lambda_{m_1,0}^{\alpha, \beta}(\theta_1, \theta_2) = 0. \\
        \end{matrix*}
    \right.
\end{equation*}

The solution to this system has the form $(\theta_1(t), \theta_2(t)) = (f_{m_1}^\alpha(t), \theta_2^0)$ for some fixed $\theta_2^0$ and $f_{m_1}^\alpha(t)$ a differentiable function depending on $m_1$ and $\alpha$ (the explicit form of $f_{m_1}^\alpha(t)$ can be obtained by solving the $1$-dimensional \ac{ODE} for $\theta_1$ by separation of variables). Thus, the flow is completely horizontal with $2m_1$ lines of critical points at the lines $\theta_1 =  \frac{2k_1 - \alpha +1}{4m_1}$, for $k_1 \in \ZZ$. Half of these critical lines are attractive, corresponding to the maxima of $f_{m_1}^\alpha$, and half of them are repulsive, corresponding to the minima. 

The situation of the Fourier modes of the form $\Lambda_{0,m_2}^{\alpha, 0}(\theta_1, \theta_2) = \sin(2\pi m_2 \theta_2)$ or $\Lambda_{0,m_2}^{\alpha, 1}(\theta_1, \theta_2) = \cos(2\pi m_2 \theta_2)$ is completely symmetric. Now, the flow is vertical and the critical lines are at $\theta_2 =  \frac{2k_2 - \alpha +1}{4m_2}$, for $k_2 \in \ZZ$ (but the attractive ones correspond to the minima and the repulsive to the minima).

Furthermore, we can collect all the Fourier modes with a vanishing frequency into a single function. To be precise, decompose the Fourier series of $\cF$ as
\begin{align*}
     \cF &= \underbrace{\frac{a_{0,0}^{0,0}}{2} + \sum_{\substack{1 \leq m_1 < \infty \\ \alpha = 0,1}} a_{m_1,0}^{\alpha,0}  \Lambda_{m_1,0}^{\alpha,0}}_{\Delta_1(\theta_1)} + \underbrace{\frac{a_{0,0}^{0,0}}{2} + \sum_{\substack{1 \leq m_2 < \infty \\ \beta = 0,1}} a_{0,m_2}^{0,\beta}  \Lambda_{0,m_2}^{0,\beta}}_{\Delta_2(\theta_2)} + \underbrace{\sum_{m_1, m_2 = 1}^\infty a_{m_1, m_2}^{\alpha,\beta}  \Lambda_{m_1,m_2}^{\alpha,\beta}}_{\Theta(\theta_1, \theta_2)}.
\end{align*}

Now, the superposition principle applied to (\ref{eq:nash-flow}) implies that any solution to the Nash flow has the form
$$
    (\theta_1(t), \theta_2(t)) = (\hat{\theta}_1(t), \theta_2^0) + (\theta_1^0, \hat{\theta}_2(t)) + \Phi(t),
$$
where $(\hat{\theta}_1(t), \theta_2^0)$ is a horizontal flow corresponding to the solution of (\ref{eq:nash-flow}) for $\Delta_1$ (explicitly, $\hat{\theta}_1$ is the solution to the equation $\hat{\theta}_1'=\frac{d}{d\theta_1} \Delta_1(\hat{\theta}_1)$), $(\theta_1^0, \hat{\theta}_2(t))$ is a vertical flow corresponding to the solution of (\ref{eq:nash-flow}) for $\Delta_2$ (i.e.\ $\hat{\theta}_2$ is the solution to $\hat{\theta}_2'=-\frac{d}{d\theta_2} \Delta_2(\hat{\theta}_2)$), and $\Phi$ is the solution to the (coupled) system of equations (\ref{eq:nash-flow}) for $\Theta$.

For this reason, in many cases the effect of the $\Delta_1$ and the $\Delta_2$ parts to the dynamics is negligible and can be ignored.

\subsection{Nash flow for Fourier basis}\label{sec:nash-flow-fourier-basis}

In this section, we shall analyze the dynamics of the Nash flow for the remaining Fourier basis. For this purpose, let us consider the function $\Lambda_{m_1, m_2}^{\alpha,\beta}$, for some $\alpha, \beta \in \ZZ_2$ with $m_1,m_2 \geq 1$. The Nash vector field associated to it is
\begin{equation}\label{eq:nash-vec-field}
    \cN\left(\Lambda_{m_1, m_2}^{\alpha,\beta}\right) =  2\pi\left((-1)^\alpha m_1 \Lambda_{m_1, m_2}^{\alpha+1,\beta}, (-1)^\beta m_2 \Lambda_{m_1, m_2}^{\alpha,\beta+1}\right).
\end{equation}

Recall that if $(\theta_1, \theta_2) \in \TT^2$ is a zero of $\Lambda_{m_1, m_2}^{\alpha,\beta}$, then it satisfies
$$
    4\theta_1m_1 \equiv 2k_1 + \alpha \mod 4\ZZ, \quad \textrm{or} \quad 4\theta_2m_2 \equiv 2k_2 + \beta \mod 4\ZZ,
$$
for some $k_1, k_2 \in \ZZ$. In other words, if we take into account the periodicity of the function $\Lambda_{m_1, m_2}^{\alpha,\beta}$, the zeros are given by
\begin{align*}
  \theta_1 = \frac{2k_1 + \alpha}{4m_1}, \quad \textrm{or} \quad
  \theta_2 = \frac{2k_2 + \beta}{4m_2},
\end{align*}
for $0 \leq k_1 < 2m_1$ and $0 \leq k_2 < 2m_2$. Observe that all these values are different, so $\Lambda_{m_1, m_2}^{\alpha,\beta}$ has $4m_1m_2$ zeros.

Coming back to Equation (\ref{eq:nash-vec-field}), we observe that if $(\theta_1, \theta_2) \in \TT^2$ is a critical point of the Nash vector field (i.e.\ a critical point of $\Lambda_{m_1, m_2}^{\alpha,\beta}$) then it satisfies one of the following two possibilities
\begin{center}
    \begin{tabular}{lll}
         \textrm{(I)} & $\displaystyle \left(4\theta_1m_1, 4\theta_2m_2\right) \equiv (2k_1 - \alpha +1 , 2k_2 - \beta + 1)$ & $\mod 4\ZZ \times 4\ZZ$, \\
        \textrm{(II)} & $\displaystyle \left(4\theta_1m_1, 4\theta_2m_2\right) \equiv (2k_1 + \alpha , 2k_2 + \beta)$ & $\mod 4\ZZ \times 4\ZZ$. \\
    \end{tabular}
\end{center}
Beware of the change of sign in the coefficient of $\alpha$ and $\beta$ for points (I). This is just a matter of notational convenience, as it will be shown below. Equivalently, the these conditions can be written explicitly as
\begin{center}
    \begin{tabular}{lll}
         \textrm{(I)} & $\displaystyle (\theta_1, \theta_2) = \left(\frac{2k_1 - \alpha + 1}{4m_1},\frac{2k_2 - \beta + 1}{4m_2}\right),
$ & for $k_1, k_2 \in \ZZ$, \\
        \textrm{(II)} & $\displaystyle (\theta_1, \theta_2) = \left(\frac{2k_1 + \alpha}{4m_1}, \frac{2k_2 + \beta}{4m_2}\right),$ & for $k_1,k_2 \in \ZZ$. \\
    \end{tabular}
\end{center}
Thus, the Nash vector field has $8m_1m_2$ critical points: $4m_1m_2$ critical points of type (I) and $4m_1m_2$ of type (II).

Regarding the Nash Hessian, it is explicitly given by
$$
    \N\Hess\left(\Lambda_{m_1, m_2}^{\alpha,\beta}\right) = 4\pi^2\begin{pmatrix}
    - m_1^2 \Lambda_{m_1, m_2}^{\alpha,\beta} & (-1)^{\alpha + \beta} m_1m_2 \Lambda_{m_1, m_2}^{\alpha+1,\beta+1}\\
    (-1)^{\alpha + \beta + 1} m_1m_2 \Lambda_{m_1, m_2}^{\alpha+1,\beta+1} & m_2^2 \Lambda_{m_1, m_2}^{\alpha,\beta}
    \end{pmatrix}
$$
Therefore, evaluated at a critical point of the form (I), we get that
\begin{align*}
    \N\Hess\left(\Lambda_{m_1, m_2}^{\alpha,\beta}\right)|_{\textrm{(I)}} = (-1)^{k_1+k_2}4\pi^2\begin{pmatrix}
    -  m_1^2  & 0\\
    0 & m_2^2 
    \end{pmatrix}.
\end{align*}
These are all saddle points for the Nash flow, with an attractive direction and a repulsive direction.

\begin{figure}[h]
	\begin{center}
	\begin{subfigure}{.47\textwidth}
	\includegraphics[scale=0.35]{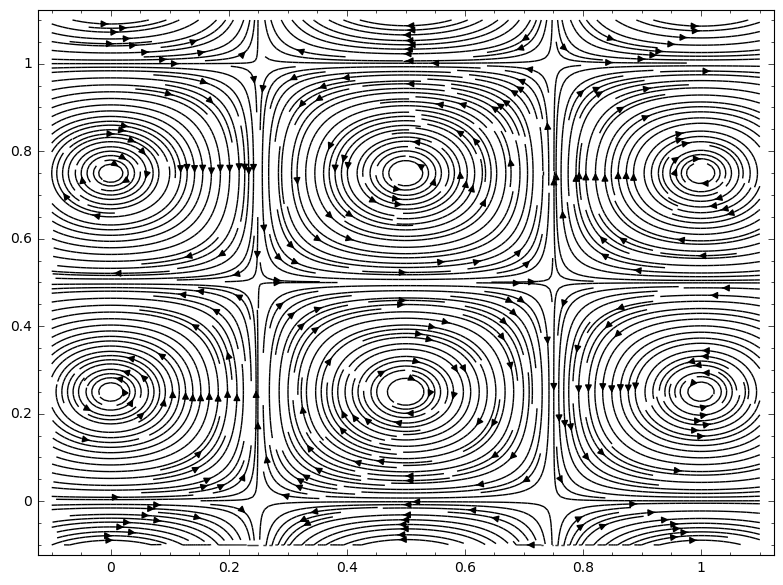}
	\subcaption[]{$\Lambda_{1,1}^{0,1} = \sin(2\pi \theta_1)\cos(2\pi \theta_2)$}
	\end{subfigure}
	\begin{subfigure}{.47\textwidth}
	\includegraphics[scale=0.35]{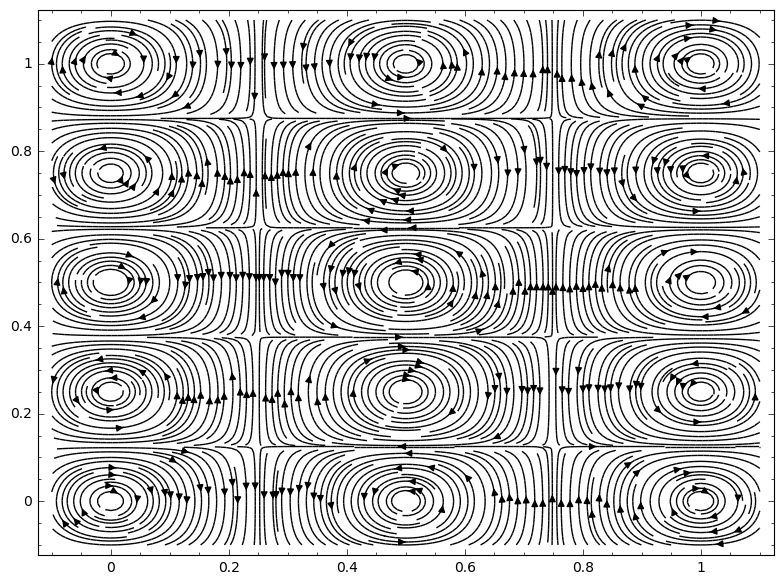}
	\subcaption[]{$\Lambda_{1,2}^{0,0} = \sin(2\pi \theta_1)\sin(4\pi \theta_2)$}
	\end{subfigure}
	
	\vspace{0.5cm}
	
	\begin{subfigure}{\textwidth}
	\begin{center}
	\includegraphics[scale=0.45]{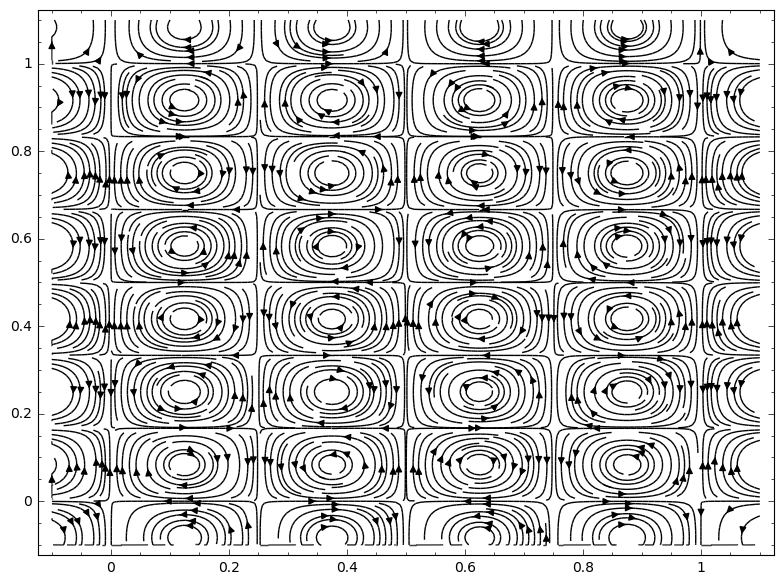}
	\subcaption[]{$\Lambda_{2,3}^{1,1}=\cos(4\pi \theta_1)\cos(6\pi \theta_2)$}
	\end{center}
	\end{subfigure}
	
	\caption{Nash flow dynamics of Fourier basis functions.}
	\label{fig:nash-flow-basis-fourier}
	\end{center}
	\end{figure}

On the other hand, the Nash Hessian evaluated at a critical point of the form (II) is
\begin{align*}
    \N\Hess\left(\Lambda_{m_1, m_2}^{\alpha,\beta}\right)|_{\textrm{(II)}}  &= (-1)^{k_1+k_2+\alpha+\beta} 4\pi^2\begin{pmatrix}
    0  & m_1m_2\\
    -m_1m_2 & 0 
    \end{pmatrix} \\
    &\sim (-1)^{k_1+k_2+\alpha+\beta} 4\pi^2 m_1m_2\begin{pmatrix}
    i  & 0\\
    0 & -i 
    \end{pmatrix}.
\end{align*}
In this situation, we get a center critical point, with periodic orbits around it and no convergent flow lines. This dynamic is depicted in Figure \ref{fig:nash-flow-basis-fourier}. Observe that, in this plot, the $2$-dimensional torus $\TT^2$ is represented as the square $[0,1] \times [0,1]$ with the boundaries identified in pairs i.e.,\ the left boundary $\left\{0\right\} \times [0,1]$ is identified with the right boundary $\left\{1\right\} \times [0,1]$ preserving the orientation, and so are the bottom boundary $[0,1] \times \left\{0\right\}$ and the upper one $\left\{1\right\} \times [0,1]$).

Putting together these calculations, we have proven the following result.

\begin{Proposition}\label{prop:dynamics-fourierbasis}
The Nash flow for the Fourier basis function $\Lambda_{m_1, m_2}^{\alpha,\beta}$ has $8m_1m_2$ critical points, whose dynamics are:
\begin{itemize}
    \item[\normalfont{(I)}] $4m_1m_2$ points are saddle points for the flow, half of them corresponding to the maxima of $\Lambda_{m_1, m_2}^{\alpha,\beta}$ and half of them to the minima.
    \item[\normalfont{(II)}] $4m_1m_2$ points are center points for the flow, surrounded by periodic orbits and corresponding to the saddle points of $\Lambda_{m_1, m_2}^{\alpha,\beta}$.
\end{itemize}
\end{Proposition}

\subsection{Nash flow for simplified truncated Fourier series}\label{sec:nash-flow-simplified}

In \cite{Mescheder-Geiger:convergence} it is proven that, under some ideal conditions, the Nash flow associated to the cost function of a \ac{GAN} has stable Nash equilibriums. For this reason, according to Proposition \ref{prop:dynamics-fourierbasis}, these cost functions cannot be basis functions of the Fourier series. In other words, its Fourier approximation (\ref{eq:Fourierapprx}) is non-trivial. Hence, in order to capture the actual dynamics of the \ac{GAN} flow, let us consider a general truncated Fourier series of the form
$$
    \Theta = \Lambda_{m_1, m_2}^{\alpha,\beta} + \mu \Lambda_{n_1, n_2}^{\gamma,\delta},
$$
for some $\alpha,\beta, \gamma, \delta \in \ZZ_2$, $-1 \leq \mu \leq 1$ and Fourier modes $m_1,m_2,n_1,n_2 \geq 1$. 

In order to simplify the computations, in this section we will suppose that $m_1=m_2=1$. After this case, the general setting will be studied. In this simplified case, at a point $(\theta_1^0, \theta_2^0)=\left({k_1/2 + \alpha/4}, {k_2/2+\beta/4}\right)$ of the form (II) we have
$$
    \nabla \Theta|_{(\theta_1^0, \theta_2^0)} = 2\pi\mu((-1)^{\gamma}n_1\Lambda_{n_1, n_2}^{\gamma+1,\delta}(\theta_1^0, \theta_2^0), (-1)^{\delta}n_2\Lambda_{n_1, n_2}^{\gamma,\delta+1}(\theta_1^0, \theta_2^0)).
$$
At this point, we have the following two options.
\begin{itemize}
    \item If $\nabla \Theta|_{(\theta_1^0, \theta_2^0)} = 0$, then $(\theta_1^0, \theta_2^0)$ is also a critical point of $\Theta$. Hence, the dynamic of the Nash flow near $(\theta_1^0, \theta_2^0)$ is determined by the Nash Hessian at that point. This Hessian is given by
    \begin{align*}
    \N\Hess\left(\Theta\right)|_{(\theta_1^0, \theta_2^0)}  = (-1)^{k_1+k_2+\alpha+\beta} 4\pi^2\begin{pmatrix}
    0  & 1\\
    -1 & 0 
    \end{pmatrix} + \mu \N\Hess\left(\Lambda_{n_1, n_2}^{\gamma,\delta}\right)|_{(\theta_1^0, \theta_2^0)}
\end{align*}

Suppose that $(\gamma, \delta) = (\alpha +1, \beta +1)$ in $\ZZ_2 \times \ZZ_2$. Set $\sigma = (-1)^{n_1k_1 + n_2k_2 + \alpha n_1/2 + \beta n_2/2}$. Observe that $\Lambda_{n_1, n_2}^{\alpha,\beta}\left(\theta_1^0, \theta_2^0\right) = 0$ and $\Lambda_{n_1, n_2}^{\alpha + 1,\beta + 1}\left(\theta_1^0, \theta_2^0\right) = \sigma$, so we have that
$$
\N\Hess\left(\Lambda_{n_1, n_2}^{\gamma,\delta}\right)|_{(\theta_1^0, \theta_2^0)} =  4\pi^2\mu\sigma\begin{pmatrix}
    - n_1^2 & 0\\
    0 & n_2^2\end{pmatrix}
$$

With this calculation at hand, we observe the following. By continuity, for $|\mu|$ small, since $\N\Hess\left(\Lambda_{1,1}^{\alpha,\beta}\right)|_{(\theta_1^0, \theta_2^0) }$ has complex eigenvalues, then $\N\Hess\left(\Theta\right)|_{(\theta_1^0, \theta_2^0)}$ also has complex eigenvalues. In particular, they must be conjugated, say $\lambda, \overline{\lambda} \in \CC$. In that case, the stability of a critical point at $(\theta_1^0, \theta_2^0)$ is governed by the trace
$$
    2\Re(\lambda) = \lambda + \overline{\lambda} = \textrm{tr} \N\Hess\left(\Theta\right)|_{(\theta_1^0, \theta_2^0)} =  4\pi^2\mu\sigma\left(n_2^2-n_1^2\right).
$$
Hence, if $n_2 < n_1$ and $\mu\sigma = 1$, or $n_2 > n_1$ and $\mu\sigma = -1$ (resp.\ $n_2 > n_1$ and $\mu\sigma = 1$, or $n_2 < n_1$ and $\mu\sigma = -1$), any critical point nearby $(\theta_1, \theta_2) \in \TT^2$ will be an spiral attractor (resp.\ repulsor). In the case that $n_1 = n_2$, the eigenvalues are multiples of $i$ and $-i$ so the point is still a center and the behaviour bifurcates depending on further Fourier modes.

On the other hand, if $\gamma = \alpha$ or $\delta = \beta$ in $\ZZ_2$, then we have that
\begin{align}\label{eq:vanishing-diag-Nash}
    \N\Hess\left(\Lambda_{n_1, n_2}^{\gamma,\delta}\right)|_{(\theta_1^0, \theta_2^0)} = \pm 4\pi^2\mu\begin{pmatrix}
    0 &  n_1n_2\\
    - n_1n_2 & 0\end{pmatrix}
\end{align}
Therefore, $\N\Hess\left(\Theta\right)|_{(\theta_1^0, \theta_2^0)}$ is still an anti-diagonal matrix and the dynamics depends of further Fourier modes.
    \item If $\nabla \Theta|_{(\theta_1^0, \theta_2^0)} \neq 0$, then $(\theta_1^0, \theta_2^0)$ is no longer a critical point of $\Theta$. However, if $|\mu|$ is small, by the implicit function theorem, nearby $(\theta_1^0, \theta_2^0)$ there must be a unique critical point $(\tilde{\theta}_1, \tilde{\theta}_2) \in \TT^2$ of $\Theta$. Again, by continuity, since $\N\Hess\left(\Lambda_{1,1}^{\alpha,\beta}\right)|_{(\theta_1, \theta_2) }$ has complex eigenvalues, then $\N\Hess\left(\Theta\right)|_{(\tilde{\theta}_1, \tilde{\theta}_2)}$ also has complex eigenvalues and their real part can be controlled through the trace.
    
    Explicitly, the Nash Hessian is
    
    \begin{align*}
    \N\Hess\left(\Theta\right)|_{(\tilde{\theta}_1, \tilde{\theta}_2)}  &= 4\pi^2\left.\begin{pmatrix}
    - n_1^2\mu \Lambda_{n_1, n_2}^{\gamma,\delta} & \pm 1 \pm \mu n_1n_2 \Lambda_{n_1, n_2}^{\gamma + 1,\delta + 1}\\
    \mp 1 \mp \mu n_1n_2 \Lambda_{n_1, n_2}^{\gamma + 1,\delta + 1} & n_2^2 \mu \Lambda_{n_1, n_2}^{\gamma ,\delta }
    \end{pmatrix}\right|_{(\tilde{\theta}_1, \tilde{\theta}_2)}
\end{align*}
and, therefore, its trace is given by
\begin{align}\label{eq:trace-simple}
    4\pi^2\mu \Lambda_{n_1, n_2}^{\gamma,\delta}(\tilde{\theta}_1, \tilde{\theta}_2) \left(n_2^2 -n_1^2\right).
\end{align}
In particular, if $n_1 = n_2$ then the new critical point $(\tilde{\theta}_1, \tilde{\theta}_2)$ is still a center. Otherwise, the behaviour is determined by the sign of $\Lambda_{n_1, n_2}^{\gamma,\delta}(\tilde{\theta}_1, \tilde{\theta}_2)$. This sign can be read from the gradient and the Nash Hessian at $(\theta_1^0, \theta_2^0)$.

To illustrate this idea, we will consider a particular combination of signs. The other cases can be obtained analogously. Suppose that the first component of the gradient satisfies
$$
\frac{\partial\Theta}{\partial \theta_1}(\theta_1^0, \theta_2^0) = 2\pi\mu(-1)^{\gamma}n_1\Lambda_{n_1, n_2}^{\gamma+1,\delta}(\theta_1^0, \theta_2^0) >0.
$$
In addition, suppose that the entries of the first row of the Nash Hessian have signs
\begin{align*}
    \left(\N\Hess\left(\Lambda_{n_1, n_2}^{\gamma,\delta}\right)|_{(\theta_1^0, \theta_2^0)}\right)_{1,1} &= - 4\pi^2n_1^2\mu \Lambda_{n_1, n_2}^{\gamma,\delta}(\theta_1^0, \theta_2^0) > 0,\\ \left(\N\Hess\left(\Lambda_{n_1, n_2}^{\gamma,\delta}\right)|_{(\theta_1^0, \theta_2^0)}\right)_{1,2} &=  \pm 1 \pm \mu n_1n_2 \Lambda_{n_1, n_2}^{\gamma + 1,\delta + 1}(\theta_1^0, \theta_2^0)<0.
\end{align*}
In that case, this means that $(\tilde{\theta}_1, \tilde{\theta}_2)$ has the form $(\tilde{\theta}_1, \tilde{\theta}_2) = (\theta_1^0-\epsilon_1, \theta_2^0+\epsilon_2)$ for small $\epsilon_1, \epsilon_2 > 0$. Therefore, the sign of (\ref{eq:trace-simple}) is determined by the sign of $\Lambda_{n_1, n_2}^{\gamma,\delta}(\theta_1^0-\epsilon_1, \theta_2^0+\epsilon_2)$, which is a well-defined quantity that only depends on the particular point $(\theta_1^0, \theta_2^0)$ and $\gamma, \delta \in \ZZ_2$.
\end{itemize}

\subsection{Nash flow for general truncated Fourier series}\label{sec:nash-flow-truncated}

In the general case, the calculation is similar but more involved. To alleviate notation, let us consider the auxiliary functions
$$
    \sigma^0(\theta) = \left\{\begin{matrix}
        0 & \textrm{if } \theta = 0 \textrm{ or } \frac{1}{2},\\
        1 & \textrm{if } 0 < \theta < \frac{1}{2}, \\
        -1 & \textrm{if } \frac{1}{2} < \theta < 1, \\
    \end{matrix}\right. \qquad \sigma^1(\theta) = \left\{\begin{matrix}
        0 & \textrm{if } \theta = \frac{1}{4} \textrm{ or } \frac{3}{4},\\
        1 & \textrm{if } 0 \leq \theta < \frac{1}{4} \textrm{ or } \frac{3}{4} \leq \theta < 1,\\
        -1 & \textrm{if } \frac{1}{4} < \theta < \frac{3}{4}. \\
    \end{matrix}\right.
$$
Notice that these maps are just the sign functions of the trigonometric functions $\sigma^0(\theta) = \textrm{sign}\left(\sin(2\pi \theta)\right)$ and $\sigma^1(\theta) = \textrm{sign}\left(\cos(2\pi \theta)\right)$, with the customary assumption that the sign function vanishes at zero. If needed, we may extend them to the whole real line by periodicity.

Now, let us consider a truncated Fourier series with arbitrary frequencies $m_1, m_2, n_1, n_2 \geq 1$ of the form
$$
    \Theta = \Lambda_{m_1, m_2}^{\alpha,\beta} + \mu \Lambda_{n_1, n_2}^{\gamma,\delta}.
$$
Analogously to the previous case, the gradient of $\Theta$ at a point 
$$
    (\theta_1^0, \theta_2^0)=\left(\frac{(2k_1 + \alpha)n_1}{4m_1},\frac{(2k_2 + \beta)n_2}{4m_2}\right) \in \TT^2
$$
of the form (II) is
$$
    \nabla \Theta|_{(\theta_1^0, \theta_2^0)} = 2\pi\mu((-1)^{\gamma}n_1\Lambda_{n_1, n_2}^{\gamma+1,\delta}(\theta_1^0, \theta_2^0), (-1)^{\delta}n_2\Lambda_{n_1, n_2}^{\gamma,\delta+1}(\theta_1^0, \theta_2^0)).
$$

Therefore, we again find a bifurcation of behaviour depending on whether $\nabla \Theta|_{(\theta_1^0, \theta_2^0)} = 0$ or not. If $\nabla \Theta|_{(\theta_1^0, \theta_2^0)} = 0$, the Nash Hessian a it is given by
\begin{align*}
    \N\Hess\left(\Theta\right)|_{(\theta_1^0, \theta_2^0)}  = (-1)^{k_1+k_2+\alpha+\beta} 4\pi^2\begin{pmatrix}
    0  & m_1m_2\\
    -m_1m_2 & 0 
    \end{pmatrix} + \mu \N\Hess\left(\Lambda_{n_1, n_2}^{\gamma,\delta}\right)|_{(\theta_1^0, \theta_2^0)}
\end{align*}

As above, the character of this matrix depends some combinatorials of $(\alpha, \beta)$ and $(\gamma, \delta)$. Explicitly, we have that

\begin{align*}
    \N\Hess\left(\Theta\right)|_{(\theta_1^0, \theta_2^0)}  &= 4\pi^2\left.\begin{pmatrix}
    - n_1^2\mu \Lambda_{n_1, n_2}^{\gamma,\delta} & \pm m_1m_2 \pm \mu n_1n_2 \Lambda_{n_1, n_2}^{\gamma + 1,\delta + 1}\\
    \mp m_1m_2 \mp \mu n_1n_2 \Lambda_{n_1, n_2}^{\gamma +1,\delta + 1} & n_2^2 \mu \Lambda_{n_1, n_2}^{\gamma,\delta}
    \end{pmatrix}\right|_{(\theta_1^0, \theta_2^0)}
\end{align*}
For $|\mu|$ small, $\N\Hess\left(\Theta\right)|_{(\theta_1^0, \theta_2^0)}$ has complex eigenvalues $\lambda, \overline{\lambda} \in \CC$. Since $\lambda + \overline{\lambda}=2\Re(\lambda)$, the dynamics are ruled by the real part $\Re(\lambda)$ which is given by the trace
$$
    4\pi^2\mu \Lambda_{n_1, n_2}^{\gamma,\delta}|_{(\theta_1^0, \theta_2^0)} \left(n_2^2 -n_1^2\right).
$$
Its negativity (resp.\ positivity) can be controlled with the trigonometric sign functions as
$$
\mu \sigma^{\gamma}(\theta_1^0n_1)\sigma^{\delta}(\theta_2^0n_2) \left(n_2^2 -n_1^2\right) <0 \textrm{     (resp. $>0$).}
$$

\begin{Remark}
There are many cases in which this trace does not vanish. For instance, if $(\gamma, \delta) = (\alpha +1, \beta +1)$ in $\ZZ_2 \times \ZZ_2$, in general,
$$\Lambda_{n_1, n_2}^{\alpha+1,\beta+1}\left(\frac{2k_1 + \alpha}{4m_1}, \frac{2k_2+\beta}{4m_2}\right) \neq 0.
$$
To be precise, given $n \in \NN$, let us denote by $\textrm{par}(n)$ the unique integer such that $n = 2^{\textrm{par}(n)}n'$ with $n'$ odd.
In that case, we have that $\Lambda_{n_1,n_2}^{\alpha+1,\beta+1}\left(\frac{2k_1 + \alpha}{4m_1}, \frac{2k_2+\beta}{4m_2}\right) = 0$ for some $k_1, k_2 \in \ZZ$ if and only if $\textrm{par}(m_1) = \textrm{par}(n_1) + (-1)^\alpha$ or $\textrm{par}(m_2) = \textrm{par}(n_2) + (-1)^\beta$. It would be interesting to study the relation between the behavior and the small divisors phenomena observed in KAM theory \cite{arnol2013mathematical}.
\end{Remark}

The case in which $\nabla \Theta|_{(\theta_1^0, \theta_2^0)} \neq 0$ can be treated similarly, but now we must not look at the Nash Hessian exactly at $(\theta_1^0, \theta_2^0)$ but at a point nearby it. Generalizing the argument of Section \ref{sec:nash-flow-simplified}, set
$$
    A =  (-1)^\gamma \mu n_1 \sigma^{\gamma +1}(\theta_1^0n_1)\sigma^{\delta}(\theta_2^0n_2),
$$
$$
    B_1 =  \mu\sigma^{\gamma}(\theta_1^0n_1)\sigma^{\delta}(\theta_2^0n_2), \quad  B_2 =(-1)^{k_1+k_2+\alpha+\beta}  m_1m_2 + (-1)^{\delta + \gamma} \mu n_1n_2 \sigma^{\gamma+1}(\theta_1^0n_1)\sigma^{\delta+1}(\theta_2^0n_2).
$$
Then, the unique critical point $(\tilde{\theta}_1, \tilde{\theta}_2)$ close to $(\theta_1^0, \theta_2^0)$ has the form
$$
    (\tilde{\theta}_1, \tilde{\theta}_2) = \left(\theta_1^0 +\textrm{sign}(AB_1)\epsilon_1, \theta_2^0+\textrm{sign}(AB_2)\epsilon_2\right),
$$
for $\epsilon_1, \epsilon_2 > 0$ small enough. Therefore, the dynamic of the critical point $(\tilde{\theta}_1, \tilde{\theta}_2)$ is determined by
\begin{align}\label{eq:check-trace}
\mu \sigma^{\gamma}((\theta_1^0 +\textrm{sign}(AB_1)\epsilon_1)n_1)\sigma^{\delta}((\theta_2^0 +\textrm{sign}(AB_2)\epsilon_2)n_2) \left(n_2^2 -n_1^2\right).
\end{align}
This quantity controls the the sign of the trace of the Nash Hessian, in analogy with the analysis of Section \ref{sec:nash-flow-simplified}. Therefore, if this last quantity is negative, then $(\tilde{\theta}_1, \tilde{\theta}_2)$ is a spiral attractor and, if it is positive, the point becomes a repulsor.

\begin{figure}[ht!]
	\begin{center}
	\begin{subfigure}{.4\textwidth}
	\includegraphics[scale=0.32]{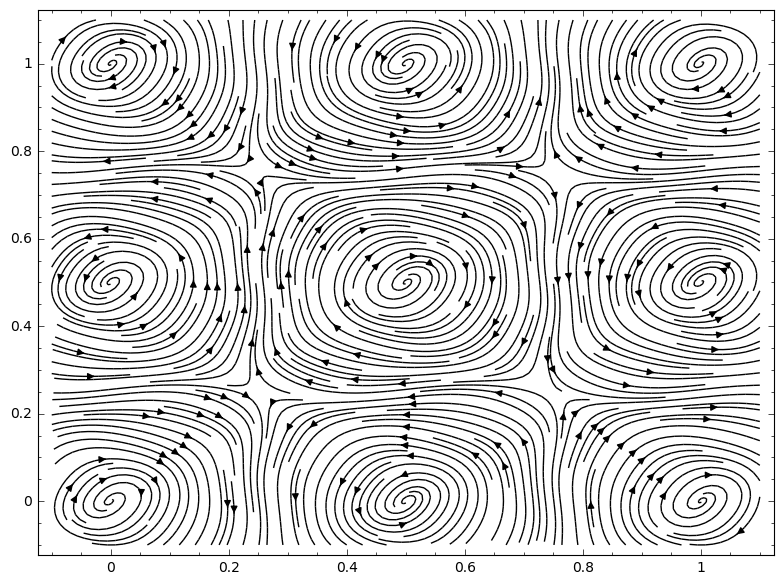}
	\subcaption[]{$\Theta = \Lambda_{1,1}^{0,0} + 0.03\Lambda_{3,5}^{1,1}$}
	\end{subfigure}
	\begin{subfigure}{.4\textwidth}
	\includegraphics[scale=0.32]{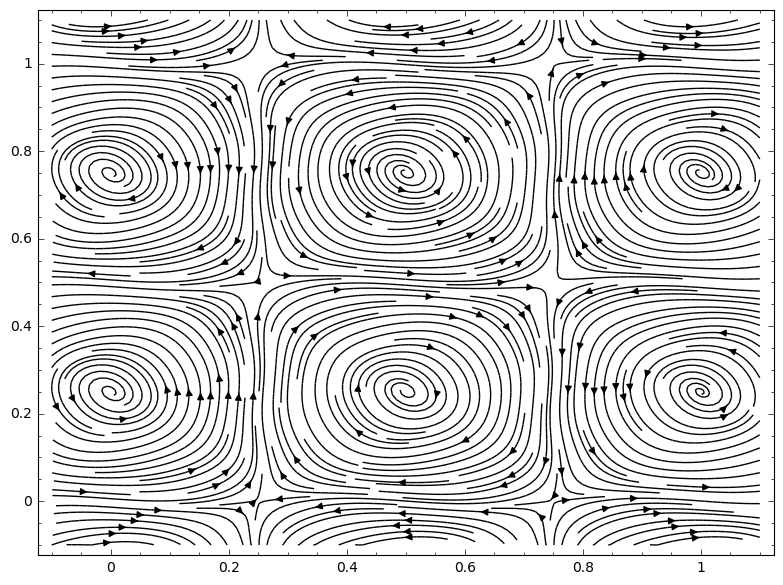}
	\subcaption[]{$\Theta = \Lambda_{1,1}^{0,1} + 0.02\Lambda_{3,5}^{1,0}$}
	\end{subfigure}
	
	\vspace{0.5cm}
	
	\begin{subfigure}{.4\textwidth}
	\includegraphics[scale=0.32]{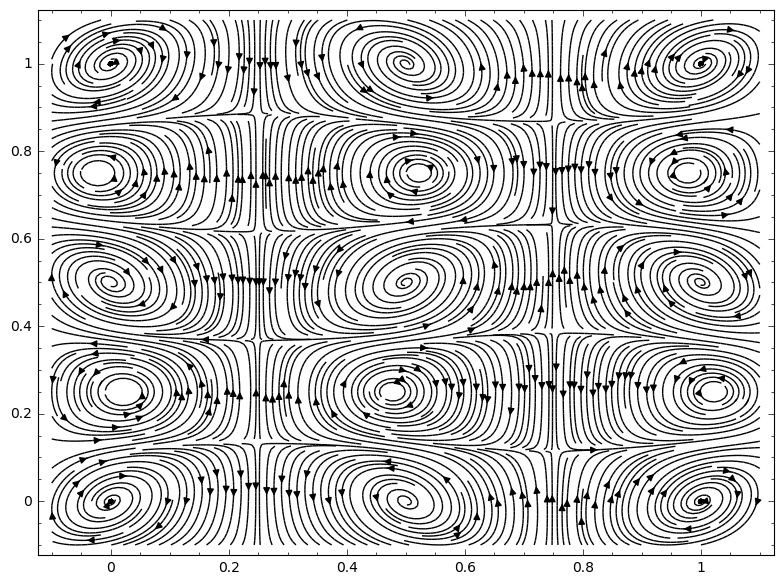}
	\subcaption[]{$\Theta = \Lambda_{1,2}^{0,0} + 0.1\Lambda_{2,3}^{1,1}$}
	\end{subfigure}
	\begin{subfigure}{.4\textwidth}
	\includegraphics[scale=0.32]{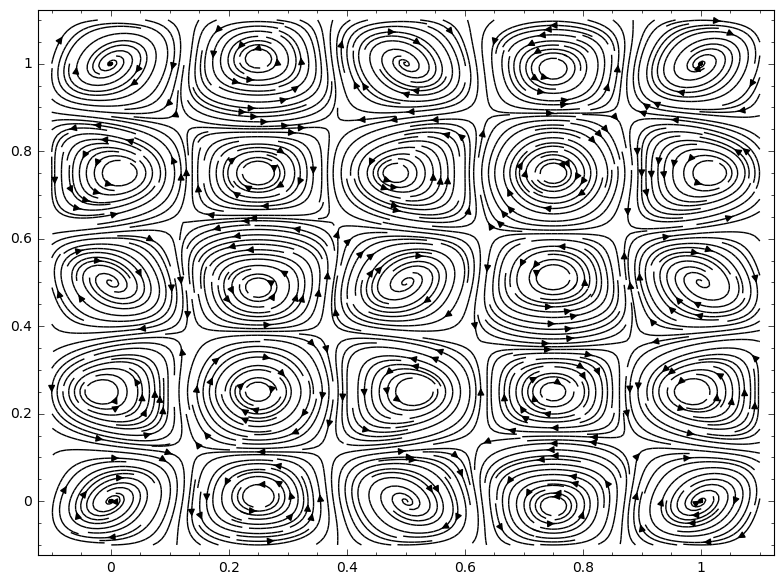}
	\subcaption[]{$\Theta = \Lambda_{2,2}^{0,0} + 0.1\Lambda_{3,5}^{1,1}$}
	\end{subfigure}

	\vspace{0.5cm}
	
	\begin{subfigure}{.4\textwidth}
	\includegraphics[scale=0.32]{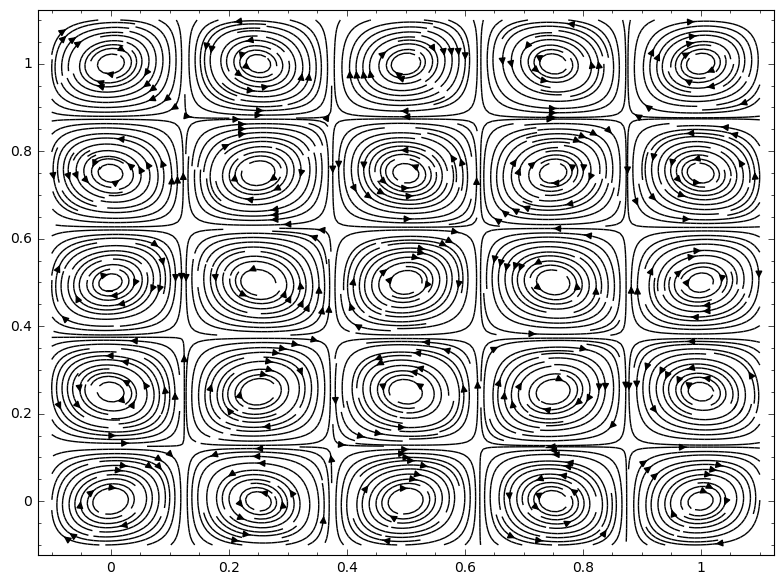}
	\subcaption[]{$\Theta = \Lambda_{2,2}^{0,0} + 0.02\Lambda_{4,4}^{1,1}$}
	\end{subfigure}
	\begin{subfigure}{.4\textwidth}
	\includegraphics[scale=0.32]{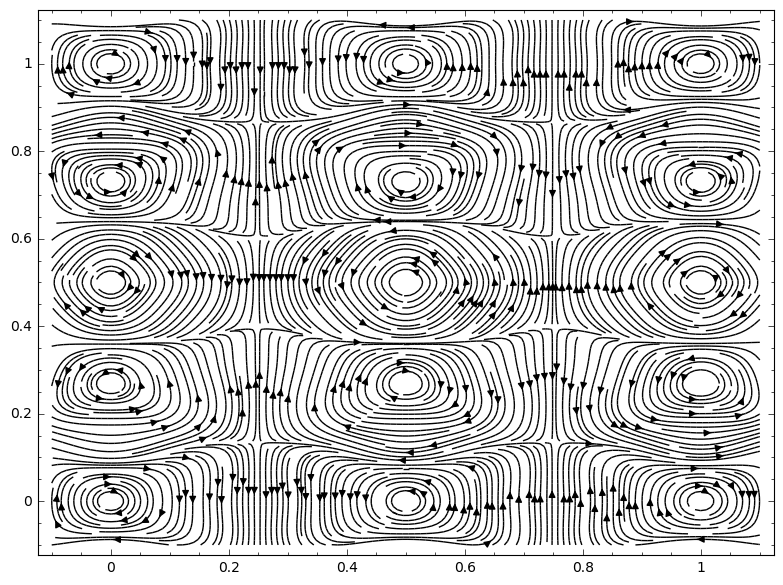}
	\subcaption[]{$\Theta = \Lambda_{1,2}^{0,0} + 0.1\Lambda_{3,5}^{0,0}$}
	\end{subfigure}
	
	\vspace{0.2cm}
	
	\caption{Nash flow dynamics of truncated Fourier series. Cases \textbf{(a)}, \textbf{(b)}, \textbf{(c)} and \textbf{(d)} show breaking of the periodic orbits into spiral flow. Cases \textbf{(e)} and \textbf{(f)} preserve the periodic orbits.}
	\label{figure:nash-flow-truncat-series}
	\end{center}
	\end{figure}

To illustrate the different bifurcation phenomena explained in this section, in Figure \ref{figure:nash-flow-truncat-series} are shown the Nash follows of some truncated series of low frequencies. Finally, summarizing this discussion, we have obtained the following result.

\begin{Theorem}
For $\mu$ small enough, the truncated Fourier series
$$
    \Theta = \Lambda_{m_1, m_2}^{\alpha,\beta} + \mu \Lambda_{n_1, n_2}^{\gamma,\delta},
$$
has an attracting (resp.\ repulsive) spiral critical point at each of the points of the form \textrm{(II)},
$$
    \displaystyle \left(\theta_1^0, \theta_2^0\right) = \left(\frac{2k_1 + \alpha}{4m_1} , \frac{2k_2 + \beta}{4m_2}\right),
$$
for $k_1, k_2 \in \ZZ$ provided that:
\begin{itemize}
        \item If $\nabla \Theta|_{(\theta_1^0, \theta_2^0)}=0$, it must hold
        $$
        \mu \sigma^{\gamma}(\theta_1^0)\sigma^{\delta}(\theta_2^0) \left(n_2^2 -n_1^2\right) <0 \textrm{     (resp. $>0$).}
        $$
        \item If $\nabla \Theta|_{(\theta_1^0, \theta_2^0)}\neq 0$, it must hold
        $$
        \mu \sigma^{\gamma}((\theta_1^0 +\textrm{sign}(AB_1)\epsilon_1)n_1)\sigma^{\delta}((\theta_2^0 +\textrm{sign}(AB_2)\epsilon_2)n_2) \left(n_2^2 -n_1^2\right) <0 \textrm{     (resp. $>0$).}
        $$
        for $\epsilon_1, \epsilon_2 > 0$ small enough.
\end{itemize}
\end{Theorem}

\begin{Remark}
Even though half of the critical points near the points of the form (II) are attractors for the Nash flow of $\Theta$, the dynamic is an small perturbation of a center. In this manner, the convergence is slow, highly spiralizing towards the Nash equilibrium. This theoretically justifies the slow and bad conditioned convergence observed in \acp{GAN} networks.
\end{Remark}

\section{Empirical analysis}\label{sec:empirical}

In this section, we show empirically how these Fourier approximations can be useful for understanding the convergence in the training of \acp{GAN}. For this purpose, in this section we will consider a simple model for a $2$-parametric torus GAN (i.e.\ with $d_D = d_G = 1$) and we shall analyze its convergence by means of its truncated Fourier series.

In the notation of Section \ref{sec:torus-gans}, we shall take $d=1$ ($1$-dimensional real data) and the parameter spaces will be $\Theta_D = \Theta_G = S^1$. The latent space will be $\Lambda = [0,1] \subseteq \RR$ with the uniform probability (standard Lebesgue measure). Fix a periodic functions $\chi: S^1 \to \RR$. Choose a $1$-parametric continuous distribution $\cD_{\xi}$ depending on the parameter $\xi \in \RR$, with cumulative distribution function $F_{\xi}$ and probability density function $f_{\xi}$. Fixed $\omega \in S^1$, the real data $X$ will be sampled according to the distribution $X \sim \cD_{\chi(\omega)}$.

As discriminator function, for $\theta_1 \in S^1$, we consider the function $D_{\theta_1}: \RR \to \RR$ given by
\begin{equation}\label{eq:discriminator}
    D_{\theta_1}(x) = \frac{f_{\chi(\omega)}(x)}{f_{\chi(\omega)}(x) + f_{\chi(\theta_1)}(x)}.
\end{equation}

On the other hand, for $\theta_2 \in S^1$, the generator will be the function $G_{\theta_2}: \Lambda = [0,1] \to \RR$ given by
\begin{equation}\label{eq:generator}
    G_{\theta_2}(\lambda) = F_{\chi(\theta_2)}^{-1}(\lambda),
\end{equation}
where $F_{\chi(\theta_2)}^{-1}$ is the quantile function of $\cD_{\chi(\theta_2)}$.

With these choices of generator and discriminator, and taking as weight function $f(t) = - \log(1 + \exp(-t))$ as in \cite{Goodfellow:2014}, the cost functional (\ref{eq:cost-fun}) reduces to
\begin{align}\label{eq:cost-fun-example}
    \mathcal{F}({\theta_1},{\theta_2}) &= \mathbb{E}_\Omega \log\left[D_{\theta_1}(X)\right] + \mathbb{E}_\Lambda \log\left[1-D_{\theta_1}(G_{\theta_2})\right] = \int_\RR \log\left(\frac{f_{\chi(\omega)}(x)}{f_{\chi(\omega)}(x) + f_{\chi(\theta_1)}(x)}\right) f_{\chi(\omega)}(x) \,dx \nonumber\\
    &+ \int_0^1 \log\left(1-\frac{f_{\chi(\omega)}\left(F_{\chi(\theta_2)}^{-1}(\lambda)\right)}{f_{\chi(\omega)}\left(F_{\chi(\theta_2)}^{-1}(\lambda)\right) + f_{\chi(\theta_1)}\left(F_{\chi(\theta_2)}^{-1}(\lambda)\right)}\right) \,d\lambda .
\end{align}

\begin{Remark}
These choices of shapes for the discriminator and generator functions are justified by \cite[Proposition 1]{Goodfellow:2014}. There, it is proven that, for fixed generator $G$ with transformed probability density function $f_G$, then the optimal discriminator $D_{\theta_1^0}$ is given by
\begin{align}\label{eq:perfect-discr}
    D_{\theta_1^0}(x) = \frac{f_{\chi(\omega)}(x)}{f_{\chi(\omega)}(x) + f_{G}(x)}.
\end{align}
On the other hand, recall that if $\Lambda = [0,1]$ with the uniform probability then $F_{\xi}^{-1}: \Lambda = [0,1] \to \RR$ is a random variable with distribution $\cD(\xi)$. Thus, in our case, $G_{\theta_2}$ is a random variable with distribution $\cD_{\chi(\theta_2)}$ and, therefore, transformed density $f_{\chi(\theta_2)}$.

In this vein, the goal of the generator $G$ given by (\ref{eq:generator}) is to adjust $\theta_2$ to reach the value $\theta_2 = \omega$, for which $G$ generates exactly the real data. At the other side, for fixed parameter $\theta_2$ for $G$, $D$ given by (\ref{eq:discriminator}) aims to tune $\theta_1$ to the value $\theta_1 = \theta_2$, for which $D$ is the perfect discriminator (\ref{eq:perfect-discr}).  
\end{Remark}

For the purposes of these experiments, we will fix as underlying distribution $\cD_\xi$ to be the exponential distribution with mean $1/\xi$, and $\chi(\theta) = \sin(\pi\theta)^2 + 1$. Recall that, in this situation, $f_\xi(x) = \xi e^{-\xi x}$ y $F_\xi(x) = 1- e^{-\xi x}$. In this way, the discriminator function (\ref{eq:discriminator}) and the generator (\ref{eq:generator}) are given by
\begin{equation}\label{eq:disc-gen-part}
    D_{\theta_1}(x) = \frac{e^{x \sin\left(\pi \theta_{1}\right)^{2}}}{\frac{\left(\sin\left(\pi \theta_{1}\right)^{2} + 1\right)}{\left(\sin\left(\pi \omega\right)^{2} + 1\right)} e^{x \sin\left(\pi \omega\right)^{2}} + e^{x \sin\left(\pi \theta_{1}\right)^{2}}}, \quad G_{\theta_2}(\lambda) = \frac{1}{\sin\left(\pi \theta_{2}\right)^{2} + 1}\log\left(-\frac{1}{\lambda - 1}\right).
\end{equation}

\begin{figure}[h]
	\begin{center}
	\begin{subfigure}{.45\textwidth}
	\includegraphics[scale=0.5]{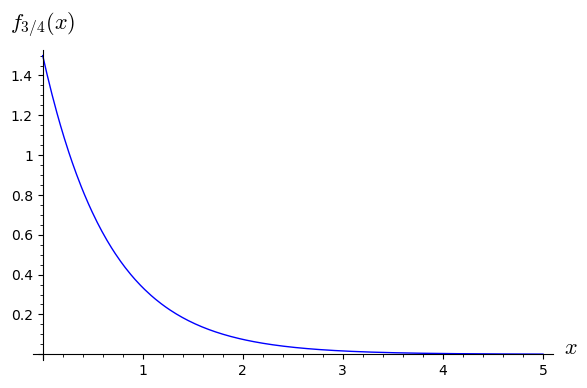}
	\subcaption[]{Probability density function}
	\end{subfigure}\hspace{1.1cm}
	\begin{subfigure}{.45\textwidth}
	\includegraphics[scale=0.5]{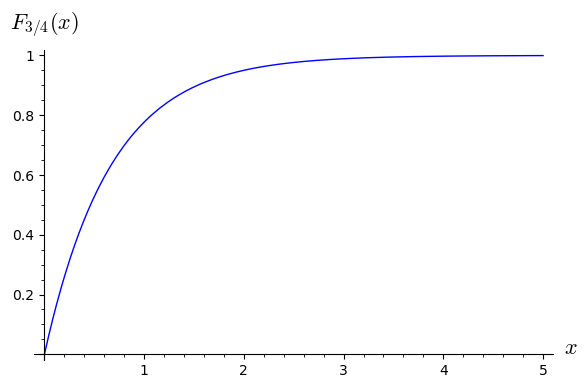}
	\subcaption[]{Cumulative distribution function}
	\end{subfigure}
	\end{center}
	\caption{Distribution of the real data}
	\label{fig:real-data}
	\end{figure}

Moreover, from now on we fix $\omega = 1/4$, so that $\chi(\omega) = 3/4$. The resulting probability density and cumulative distribution functions of the real data are plotted in Figure \ref{fig:real-data}.

\begin{figure}[h]
	\begin{center}
	\begin{subfigure}{.45\textwidth}
	\includegraphics[scale=0.5]{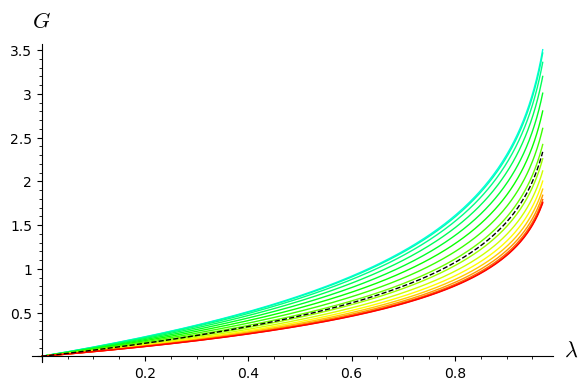}
	\subcaption[]{Output of the function}
	\end{subfigure}\hspace{1.1cm}
	\begin{subfigure}{.45\textwidth}
	\includegraphics[scale=0.5]{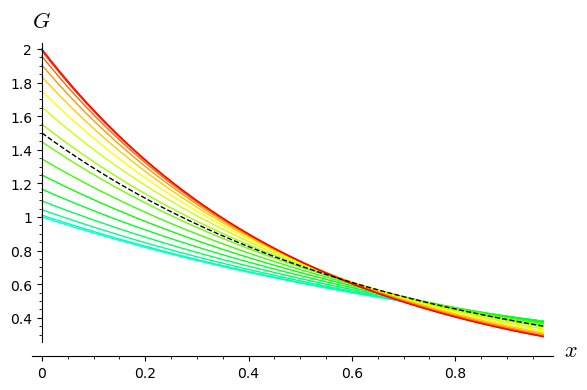}
	\subcaption[]{Transformed probability density function}
	\end{subfigure}
	\end{center}
	\caption{Generator functions for $0 \leq \theta_2 \leq \frac{1}{2}$. The warmer the plot, the bigger the value of $\theta_2$. The dashed line corresponds to the real data.}
	\label{fig:generator}
	\end{figure}

\begin{figure}[h]
	\begin{center}
	\includegraphics[scale=0.6]{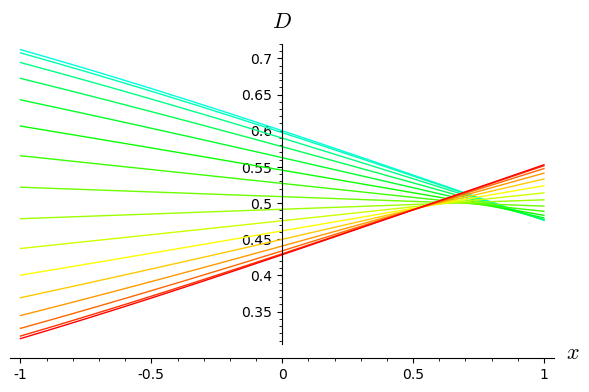}
	\end{center}
	\caption{Discriminator functions for $0 \leq \theta_1 \leq \frac{1}{2}$. The warmer the plot, the bigger the value of $\theta_1$. For fixed generator parameter $\theta_2$, the optimal value for $\theta_1$ is corresponds to the line with $\theta_1 =\theta_2$.}
	\label{fig:discriminator-fun}
	\end{figure}

With this choice of real distribution, the generator function, as well as the transformed probability density function are plotted in Figure \ref{fig:generator}, and the discriminator function is shown in Figure \ref{fig:discriminator-fun}.

\begin{figure}[h]
	\begin{center}
	\begin{subfigure}{.55\textwidth}
	\vspace{0.4cm}
	\includegraphics[scale=0.45]{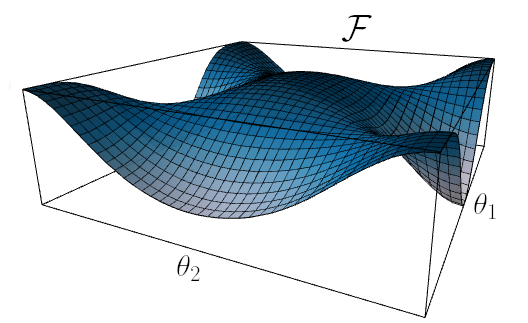}
	\vspace{0.4cm}
	\caption{Plot of the function $\cF(\theta_1, \theta_2)$. The four saddle points lie nearby each of the four corners of the frame.}
	\end{subfigure}
	\begin{subfigure}{.35\textwidth}
	\vspace{0.6cm}
	\includegraphics[scale=0.5]{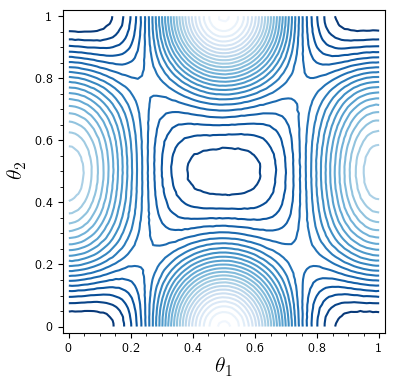}
	\caption{Contour plot of $\cF(\theta_1, \theta_2)$.}
	\end{subfigure}
	\end{center}
	\caption{Graphical representation of the landscape of the cost function $\cF(\theta_1, \theta_2): \TT^2 \to \RR$.}
	\label{fig:cost-function}
	\end{figure}

In addition, in Figure \ref{fig:cost-function} we show graphically the cost function $\cF(\theta_1, \theta_2)$ of (\ref{eq:cost-fun-example}) on $\TT^2$. The numerical approximation of the integrals in (\ref{eq:cost-fun-example}) have been carried out with the Simpson rule. The function was sampled at $225$ knot points and subsequently interpolated by means of a multiquadratic radial basis interpolation. Observe that one of the Nash equilibria of $\cF$ is at $(\theta_1, \theta_2)=(1/4,1/4)$ (bottom corner of the plot). Moreover, by the symmetries of $\chi$, the plot suggests that  $(\theta_1, \theta_2)=(1/4,3/4), (3/4,1/4), (3/4,3/4)$ are also Nash equilibria.

\begin{figure}[h]
	\begin{center}
	\includegraphics[scale=0.75]{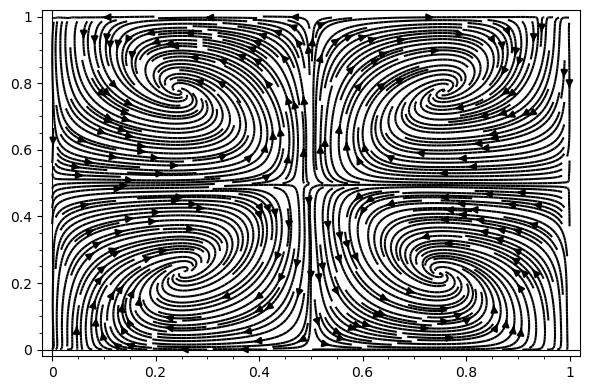}
	\end{center}
	\caption{Dynamics of the Nash flow for the torus \ac{GAN}. Four attractive Nash equilibria can be observed.}
	\label{fig:nash-flow}
	\end{figure}

In Figure \ref{fig:nash-flow}, we show the Nash flow associated to the cost function $\cF: \TT^2 \to \RR$. As it can be checked in the image, the flow confirms that there exists four Nash equilibrium points, corresponding to $(\theta_1^0, \theta_2^0)=(1/4,1/4), (1/4,3/4), (3/4,1/4)$ and $(3/4,3/4)$, all of them being attractors for the Nash flow. Another four critical points of $\cF$ can be observed in the figure: the points $(0,0)$ and $(1/2,1/2)$ correspond to the two maxima of $\cF$, and the points $(0,1/2)$ and $(1/2,0)$ to the two minima. Observe that these critical points are saddle points for the flow, with an attractive direction and a repulsive direction. Finally, notice that (\ref{eq:crit-Euler}) is satisfied since the maxima and minima have even indices ($2$ and $0$, respectively), and the Nash equilibria have odd indices.

Now, let us decompose $\cF$ according to its Fourier series. In Table \ref{tab:fourier-modes} we show the modes with the largest absolute Fourier coefficients. These coefficients have been computed using the formulae of Section \ref{sec:dynamics-fourier}, by applying rectangular quadrature as numerical integration method and looking at the modes with $1 \leq m_1, m_2 \leq 10$.

\begin{table}[]
    \centering
    \begin{tabular}{c|c|c|c|c|c}
        $m_1$ & $m_2$ & $\alpha$ & $\beta$ & $a_{m_1, m_2}^{\alpha, \beta}$ & Ratio \\\hline
 1 & 1  & 1 & 1 & 0.06127 & 1.0000\\\hline
 1 & 2 & 1 & 1 & 0.01102 & 0.1800 \\\hline
 2 & 1 & 1 & 1 & -0.00503 & -0.0822\\\hline
 2 & 2  & 1 & 1 & -0.00404 &  -0.0660 \\\hline
 2 & 3  & 1 & 1 & -0.00325 & -0.0532 \\\hline
 2 & 4  & 1 & 1 & -0.00308 & -0.0504 \\\hline
 2 & 5  & 1 & 1 & -0.00305 & -0.0499 \\\hline
 2 & 7  & 1 & 1 & -0.00304 & -0.0497 \\\hline
 2 & 9  & 1 & 1 & -0.00304 & -0.0496 \\\hline
 2 & 10 & 1 & 1 & -0.00304 & -0.0496  \end{tabular}
    \caption{Fourier modes of the cost function for the torus \ac{GAN}. The ten modes with the largest absolute value of their associated coefficient are shown. The last column shows the ratio between each Fourier coefficient and the largest coefficient.}
    \label{tab:fourier-modes}
\end{table}

From these results, we observe that the predominant Fourier modes of $\cF$ are cosine basis functions, $\Lambda_{m_1,m_2}^{1,1}(\theta_1, \theta_2) = \cos(2\pi m_1\theta_1)\cos(2\pi m_2\theta_2)$. The largest coefficient correspond to the mode $(m_1, m_2) = (1,1)$. Observe that this is not surprising: $(m_1, m_2) = (1,1)$ is the unique mode with four critical points of type (II), which correspond to the four Nash equilibria of Figure \ref{fig:nash-flow} (in other words, the four saddle points in Figure \ref{fig:cost-function}). 

For $s \geq 0$, let us order the first $s$ Fourier modes decreasingly according to the absolute value of their coefficient, $(m_1^0,m_2^0)=(1,1),(m_1^1,m_2^1), \ldots, (m_1^s,m_2^s)$. Denote by $b_{m_i^i,m_2^i}^{1,1} = a_{m_i^i,m_2^i}^{1,1}/ a_{m_i^0,m_2^0}^{1,1}$ the ratio of the Fourier coefficients. We can approximate the Nash flow of the cost function $\cF$ by the truncated Fourier series
\begin{align*}
    \Theta_s(\theta_1, \theta_2) = \Lambda_{m_1^0,m_2^0}^{1,1}(\theta_1, \theta_2) + \sum_{i=1}^s b_{m_i^i,m_2^i}^{1,1} \Lambda_{m_i^i,m_2^i}^{1,1} (\theta_1, \theta_2).
\end{align*}

\begin{figure}[ht!]
	\begin{center}
	\begin{subfigure}{.45\textwidth}
	\includegraphics[scale=0.47]{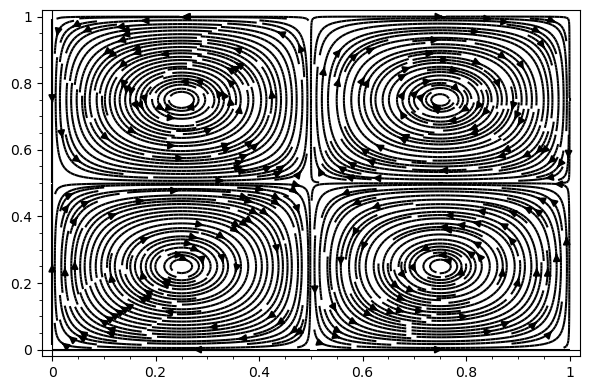}
	\subcaption[]{Approximation $\Theta_0$}
	\end{subfigure}
	\begin{subfigure}{.45\textwidth}
	\includegraphics[scale=0.47]{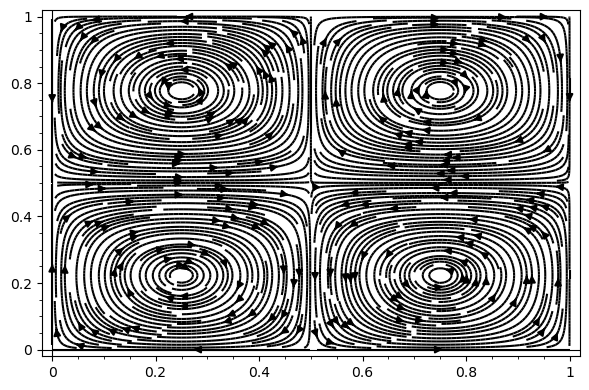}
	\subcaption[]{Approximation $\Theta_1$}
	\end{subfigure}
	
	\vspace{0.3cm}
	
	\begin{subfigure}{.45\textwidth}
	\includegraphics[scale=0.47]{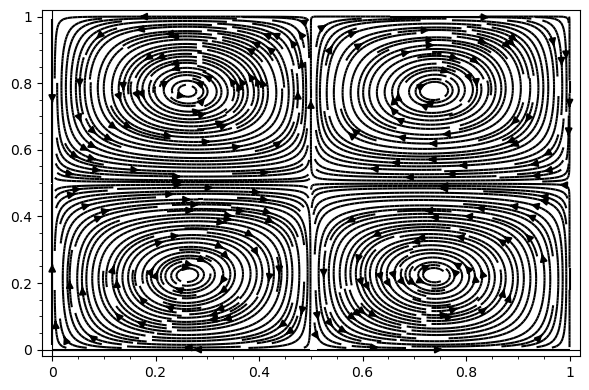}
	\subcaption[]{Approximation $\Theta_2$}
	\end{subfigure}
	\begin{subfigure}{.45\textwidth}
	\includegraphics[scale=0.47]{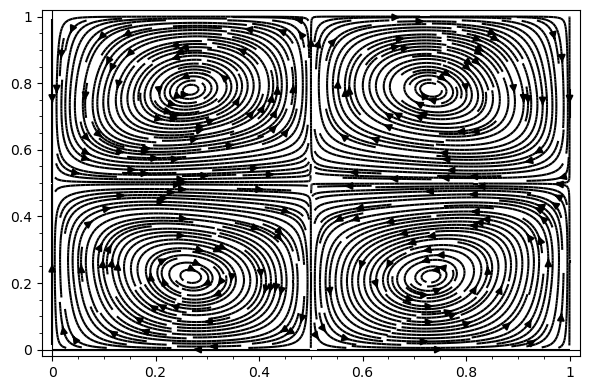}
	\subcaption[]{Approximation $\Theta_3$}
	\end{subfigure}

	\vspace{0.3cm}
	
	\begin{subfigure}{\textwidth}
	\begin{center}
	\includegraphics[scale=0.7]{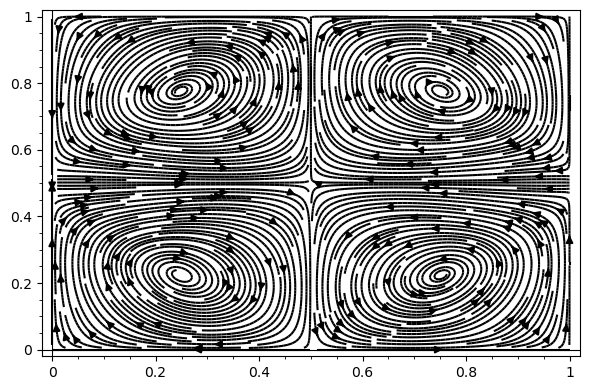}
	\subcaption[]{Approximation $\Theta_4$}
	\label{fig:nash-flow-approx-important}
	\end{center}
	\end{subfigure}
	\caption{Nash flow dynamics of truncated Fourier series approximations for the cost function of the torus \ac{GAN}.}
	\label{fig:nash-flow-approx}
	\end{center}
	\end{figure}

The associated Nash flow is depicted in Figure \ref{fig:nash-flow-approx}. As can be checked there, the critical points nearby points of type (II) are (approximately) centers for $s \leq 3$. The reason for this behavior is twofold. In the following, let $(\theta_1^0, \theta_2^0)=(1/4,1/4), (1/4,3/4), (3/4,1/4)$ or $(3/4,3/4)$.
\begin{itemize}
    \item For $s \leq 2$, we have that $\nabla \Theta_s|_{(\theta_1^0, \theta_2^0)} = 0$ since, in the gradient, there is always a term with a factor $\cos(2\pi \theta)$ that vanishes at these points. Hence, the critical point of $\Theta_s$ is exactly at $(\theta_1^0, \theta_2^0)$. Nevertheless, since all the terms $\Lambda_{m_1, m_2}^{\alpha, \beta}$ appearing in the Fourier series have equal $(\alpha, \beta)=1$, as mentioned in Section \ref{sec:nash-flow-simplified} we still have that the Nash Hessian has the form (\ref{eq:vanishing-diag-Nash}) with vanishing diagonal entries. Hence, the critical point $(\theta_1^0, \theta_2^0)$ is still a center.
    \item For $s = 3$, we find that $\nabla \Theta_3|_{(\theta_1^0, \theta_2^0)} \neq 0$ so a new critical point $(\tilde{\theta}_1, \tilde{\theta}_2)$ appears near $(\theta_1^0, \theta_2^0)$. Nevertheless, for this new mode we have that $m_1^3=m_2^3 = 2$ so Equation (\ref{eq:check-trace}) still vanishes, proving that the new critical point is still a center.
\end{itemize}
    
Finally, let us consider the case $s = 4$. In this situation, we also have $\nabla \Theta_4|_{(\theta_1^0, \theta_2^0)} \neq 0$ so a new critical point $(\tilde{\theta}_1, \tilde{\theta}_2)$ appears near $(\theta_1^0, \theta_2^0)$. The dynamic around it is governed by Equation (\ref{eq:check-trace}). To do so, we calculate the sign of the quantities $A, B_1$ and $B_2$ of Section \ref{sec:nash-flow-truncated} and we get
    $$
        A > 0, \quad B_1 < 0, \quad B_2 <0.
    $$
    Hence, the new critical point has the form $(\tilde{\theta}_1, \tilde{\theta}_2) = (\theta_1^0 - \epsilon_1, \theta_2^0- \epsilon_2)$ for $\epsilon_1, \epsilon_2 >0$ small. For these values, we have that
    $$
    \sigma^{1}(2(\theta_1^0 - \epsilon_1)) = -1, \quad \sigma^{\delta}(3(\theta_2^0 -\epsilon_2)) = -1.
    $$
    Therefore, checking Equation (\ref{eq:check-trace}) we get
    $$
     \mu\sigma^{1}(n_1(\theta_1^0 - \epsilon_1))\sigma^{\delta}(n_2(\theta_2^0 -\epsilon_2))\cdot \left(n_2^2 -n_1^2\right) = -0.003\cdot(-1) \cdot(-1)(3^2-2^2) < 0.
    $$

Therefore, for $s = 4$, the trend changes and the centers turn into spiral attractor critical points. This is the attractive behavior observed in Figure \ref{fig:nash-flow-approx-important}. Notice that this dynamic agrees with the real one observed in Figure \ref{fig:nash-flow}, which empirically confirms the validity of our approach.

\section{Methodology for practical applications}
\label{sec:methodology}

The discussion of Sections \ref{sec:dynamics-fourier} and \ref{sec:empirical} opens the door to a practical application of the analysis techniques introduced in this paper to study convergence of real-world \acp{GAN}. Observe that, in general, the knowledge of the underlying cost function $\cF$ (c.f.\ Equation (\ref{eq:cost-fun})) of a \ac{GAN} is very limited. Indeed, several metrics have been proposed in the literature to screen the evolution of the training of the \ac{GAN}. These metrics provide a way of measuring indirectly the convergence of the \ac{GAN}, but definitely skip a thorough analysis of the cost function.
Nevertheless, using the techniques introduced in this paper, we will show that it is possible to methodically analyze the dynamics of the Nash flow for the \ac{GAN} problem through the partial sums of the Fourier series of the cost function. It is remarkable that this valuable information about the behaviour of the training process cannot be extracted from $\cF$ itself. 

In this section, we aim to organize the previous analysis into a precise methodology that can be applied in practice. As it will become clear, this process was implicit in the reasoning provided in Section \ref{sec:empirical}. The proposed process of analysis comprises the following steps:

\begin{enumerate}
    \item Evaluate cost function $\cF(\theta_D, \theta_G)$ in a uniform grid for the parameters $(\theta_D, \theta_G)$ (the weights of the two neural networks forming the \ac{GAN} in the deep learning framework). Observe that for these evaluations it is not necessary to train the \ac{GAN} networks. The sampling process amounts to fixing the weights of the networks and to compute the mean prediction error of the discriminant against real and synthetic instances. No optimization of the weights must be carried out.
    \item Compute the \ac{DFT} of $\cF$ by means of the obtained samples. This process can be done efficiently through the \ac{FFT} algorithm. 
    \item Use the results of the \ac{DFT} to estimate the Fourier modes and coefficients of $\cF$. Sort the modes decreasingly according to the absolute value of their associated Fourier coefficient. 
    \item\label{step:analysis} Consider a truncation level $s \geq 0$ (starting with $s=0$). Compute the critical points of $\Theta_s$, the truncated Fourier series of $\cF$ with $s$ terms. Using the techniques developed in Section \ref{sec:dynamics-fourier} (see also Section \ref{sec:empirical}), analyze the local dynamics of the Nash flow around the critical points of $\Theta_s$.
    \item\label{step:increase} While some of the critical points of $\Theta_s$ are a center, increase the truncation level by $1$. Repeat the steps \ref{step:analysis}--\ref{step:increase} until reaching a truncation level $s_0$ such that all the critical points of $\Theta_{s_0}$ are either attractors or repulsors.
\end{enumerate}

After this process, we will have found a truncation level $s_0$ such that the local dynamics of $\Theta_{s_0}$ around the critical points are conjugated to the local dynamics of $\cF$ around its Nash equilibria. This information can be exploited to analyze the training process of the \ac{GAN}. For instance, if the convergence to the critical point is very slow, in the sense that the trace of the Nash Hessian is close to zero, then a hard convergence of the training process should be expected. This will lead to remarkable unstabilities during the learning process that may prevent the system to converge with a raw gradient descent optimization procedure. In that case, the obtained results strongly suggest that several heuristics for stabilizing the training process must be implemented. Additionally, since the equilibria are spiral attractors, if the learning rate of the gradient descend method is not small enough, the discrete time approximation may not converge. In that case, the information about the convergence rate in the simplified Fourier model can be used to properly anneal the learning rate, leading to a much stable convergence.

Despite the utility of the proposed methodology, it suffers several issues that must be addressed in future works to obtain an efficient analysis procedure. The first one is that the previous proposal has an obvious bottleneck: the sampling process of the cost function on the parameters $(\theta_D, \theta_G)$ may require a huge number of samples due to the course of dimensionality. Nevertheless, it is important to mention that it is not necessary to use a very dense grid since we want to understand the Fourier modes of the cost function $\cF$ and not to obtain a detailed picture of the landscape of $\cF$. This will largely alleviate the sampling process to make it feasible.

Another possible solution is to not sample on the whole $(\theta_D, \theta_G)$-space, but on a smaller dimensional subspace concentrating the flow. For that purpose, the \ac{GAN} network can be trained and, after some epochs, flow will have entered in a certain `convergence subspace' that will enclose the long-time evolution of the flow. This subspace can be estimated by several methods, for instance by considering the subspace generated by the last $k \geq 1$ gradient vectors obtained in the training process. In that case, instead of working on the high dimensional $(\theta_D, \theta_G)$-space, we can restrict our analysis to the $k$-dimensional affine space generated by these vectors. This is a much smaller subspace in which the sampling process can be carried out. Nevertheless, proposing other efficient methods of sampling that enable accurate approximations of the Fourier series of $\cF$ is an interesting topic for future work.
 
Another important remark is that the methodology proposed to estimate the Fourier series through the \ac{FFT} is much more efficient than the quadrature methods used in Section \ref{sec:empirical}. However, it also may lead to poorer estimations of the Fourier coefficients. This inaccuracy may produce errors when choosing the leading Fourier modes if their importance (absolute value of their Fourier coefficients) are similar. To avoid these problems, all the possible permutations of these similar modes (say, modes whose coefficients differ less that a fixed threshold) must be considered during the analysis of Nash flow of the Fourier series.
 
\section{Conclusions}\label{sec:conclusions}
 
In this paper we have studied a novel approach to deeply analyze the converge of \ac{GAN} networks on tori. This is an outstanding open problem in Machine Learning and Deep Learning that prevents \acp{GAN} to be suitable for use in arbitrary domains, as feature generation outside the world of image processing.

In this paper, we propose to decompose the cost function of a \ac{GAN} into its Fourier mode and to envisage the dynamics around the Nash equilibria through its truncated Fourier approximation. For that purpose, we have performed a thorough analysis of the dynamics of trigonometric series with one and two terms. Roughly speaking, this analysis has shown that if we truncate the Fourier series at its first mode, all the critical points are centers surrounded by periodic orbits. When we add subtler Fourier modes to the approximation, this dynamic may be preserved or may bifurcate to give rise to spiral attractors or repulsors. This dynamic is essentially determined by the trace of the Nash Hessian of the cost function. Hence, following this idea, in this paper we have exhibit explicitly the bifurcation condition for the Nash flow of the truncated Fourier approximations. These conditions have an involved shape taking into account the monotonicity of the trigonometric functions on a neighborhood of the critical point but, eventually, the conditions are very explicit and can be easily checked. As byproduct of this analysis, we have observed that, even though the Nash equilibria are stable points as proven in \cite{Mescheder-Geiger:convergence}, the dynamic of the training process is close to a center and the convergence is slow and spiral.

To test this idea, we have conducted an experimental analysis with a torus \ac{GAN} toy-model. Through this example, we have observed that the number and distribution of the critical points is determined by the first Fourier model. Nevertheless, it was necessary to reach the forth Fourier term to discover the attractive dynamics, as predicted in the \ac{GAN} literature. Comparing the approximated flow with the real flow, we observe that the approximation is able to replicate not only the local but also the global dynamics of real \ac{GAN}.

We expect that this work will be useful for quantifying the complexity and convergence properties of \ac{GAN}. To show how this theoretical analysis can be put into practice, in Section \ref{sec:methodology} we propose a methodology of analysis that enables a characterization of the training dynamics of real-world \acp{GAN} by means of the techniques developed in this work. From the obtained information about the convergence of the learning process of the networks, several improvements for stabilizing the training can be implemented, like a progressive reduction of the learning rate to adapt the geometry of the spiral flow.

It is worth mentioning that the results presented in this paper do not only apply to torus toy-models, but also to more realistic networks. It may seem at a first sight that standard \acp{GAN} do not fulfil the periodicity requirement to be defined on a torus. However, in many cases, the outputs of the generator and the discriminator networks are clipped for large enough inputs. This fix is crucial to maintain several required analytic properties, as the Lipschitz condition for Wasserstein \acp{GAN} \cite{Arjovsky-WGAN}. After this clipping, the \ac{GAN} does actually turn into a torus \ac{GAN} since the generator and discriminator functions are periodic (with a large period). In this manner, most of the regular \acp{GAN} used in image generation and feature generation fit in the framework introduced in this paper. And this is crucial, since dynamics on a closed manifold are deeply related to the underlying topology, for instance through the Poincar\'e-Hopf theorem or deeper Morse-like results.

Nevertheless, much work must be done before this project can be turned into a reality. First, in order to compute the Fourier series of the cost function, we had to sample the cost function of the \ac{GAN} at a dense mesh of weights. Using this sampling, we were able to estimate the Fourier coefficients through standard quadrature techniques, as the Simpson rule. In shallow networks with few neurons a similar approach can be applied, but for deeper networks this dense sampling is unfeasible. For this reason, better methods for estimating the Fourier coefficients of the cost function are needed, maybe by exploding the analytical and harmonical properties of the trigonometric functions. In addition, to illustrate the method, in this paper we have carried out all the calculations on a $2$-dimensional torus. The computation in higher dimensional tori may follow similar lines, but definitely a thorough analysis of the bifurcation conditions in the higher dimensional setting is not obvious.

Summarizing, in this paper we have introduced a novel method for understanding the dynamics of \acp{GAN} through harmonic analysis. We have shown that, despite that the Nash equilibria of the \ac{GAN} are stable, the convergence is a perturbation of a center and, thus, slow and complicated. The method has allow us to identify a simplified model of the dynamics that may be useful for tuning several hyper-parameters of the used \acp{GAN}, as the learning rate of the number of epochs to be trained. We expect that this work will open the door to new methods of study of dynamics of \ac{GAN} by using harmonic analysis and trascendental methods.

\vspace{6pt} 

\bibliography{bibliography.bib}{}
\bibliographystyle{abbrv}

\end{document}